\def\year{2021}\relax
\documentclass[letterpaper]{article} 
\usepackage{aaai21}  
\usepackage{times}  
\usepackage{helvet} 
\usepackage{courier}  
\usepackage[hyphens]{url}  
\usepackage[ruled,vlined]{algorithm2e}
\usepackage{xr}
\usepackage{url}
\usepackage{array}
\usepackage{cellspace}
\usepackage{makecell}
\usepackage[switch]{lineno}

\usepackage{multirow}
\usepackage{hhline}
\usepackage{caption}
\usepackage{subcaption}
\usepackage{mathtools}
\usepackage{graphicx} 
\usepackage{natbib}  
\usepackage{caption} 
\usepackage{custom_tex}
\usepackage{color}
\usepackage{outlines}
\usepackage{enumitem}

\makeatletter
\newcommand*{\addFileDependency}[1]{
  \typeout{(#1)}
  \@addtofilelist{#1}
  \IfFileExists{#1}{}{\typeout{No file #1.}}
}
\makeatother

\newcommand*{\myexternaldocument}[1]{%
    \externaldocument{#1}%
    \addFileDependency{#1.tex}%
    \addFileDependency{#1.aux}%
}

\myexternaldocument{Appendix_main}

\DeclareMathOperator*{\argmax}{arg\,max}

\newcommand{\jni}[1]{\textcolor{blue}{#1}}

\usepackage{mathrsfs}

\title{Dynamic Gaussian Mixture based Deep Generative Model For\protect\\Robust Forecasting on Sparse Multivariate Time Series}
\author{Paper ID: 1069}

\title{Dynamic Gaussian Mixture based Deep Generative Model For\protect\\Robust Forecasting on Sparse Multivariate Time Series}
\author{Yinjun Wu\textsuperscript{\rm 1}\thanks{This work was done when the first author was an intern at NEC Laboratories America. {$^\ddagger$}Corresponding author.},
Jingchao Ni\textsuperscript{\rm 2}{$^\ddagger$},
Wei Cheng\textsuperscript{\rm 2},
Bo Zong\textsuperscript{\rm 2},
Dongjin Song\textsuperscript{\rm 3},
Zhengzhang Chen\textsuperscript{\rm 2},\\
Yanchi Liu\textsuperscript{\rm 2},
Xuchao Zhang\textsuperscript{\rm 2},
Haifeng Chen\textsuperscript{\rm 2},
Susan Davidson\textsuperscript{\rm 1}\\}
\affiliations {
\textsuperscript{\rm 1}University of Pennsylvania,
\textsuperscript{\rm 2}NEC Laboratories America,
\textsuperscript{\rm 3}University of Connecticut\\
\textsuperscript{\rm 1}\{wuyinjun@seas.upenn.edu, susan@cis.upenn.edu\}\\
\textsuperscript{\rm 2}\{jni, weicheng, bzong, zchen, yanchi, xuczhang, haifeng\}@nec-labs.com,
\textsuperscript{\rm 3}dongjin.song@uconn.edu}

\begin{document}
\maketitle

\begin{abstract}
Forecasting on sparse multivariate time series (MTS) aims to model the predictors of future values of time series given their 
incomplete past, which is important for many emerging applications. However, most existing methods process MTS's individually, and do not leverage 
the dynamic distributions underlying 
the MTS's, 
leading to sub-optimal results when the sparsity is high. To address this challenge, we propose a novel 
generative model, which tracks the transition of latent clusters, instead of isolated feature representations, to achieve robust modeling. It is characterized by a newly designed dynamic Gaussian mixture distribution, which captures the dynamics of clustering structures, and is used for emitting time series. 
The generative model is parameterized by neural networks. A structured inference network is also designed for enabling inductive analysis.
A gating mechanism is further introduced to dynamically tune the Gaussian mixture distributions. Extensive experimental results on a variety of real-life datasets 
demonstrate the effectiveness of our method.
\end{abstract}
\section{Introduction}

Multivariate time series (MTS) analysis is heavily used in a variety of applications, such as cyber-physical system monitoring \cite{zhang2019deep}, financial forecasting \cite{binkowski2018autoregressive}, traffic analysis \cite{li2018diffusion}, and clinical diagnosis \cite{che2018recurrent}. 
Recent advances in deep learning have spurred on many innovative machine learning models on MTS data, 
which have shown remarkable results on a number of fundamental tasks, including forecasting \cite{qin2017dual}, event prediction \cite{choi2016retain}, and anomaly detection \cite{zhang2019deep}. Despite these successes, most  existing models treat the input MTS as homogeneous and complete sequences. In many emerging applications, however, MTS signals are integrated from heterogeneous sources and are very sparse.


\begin{figure}[!t]
\includegraphics[width=1.0\columnwidth]{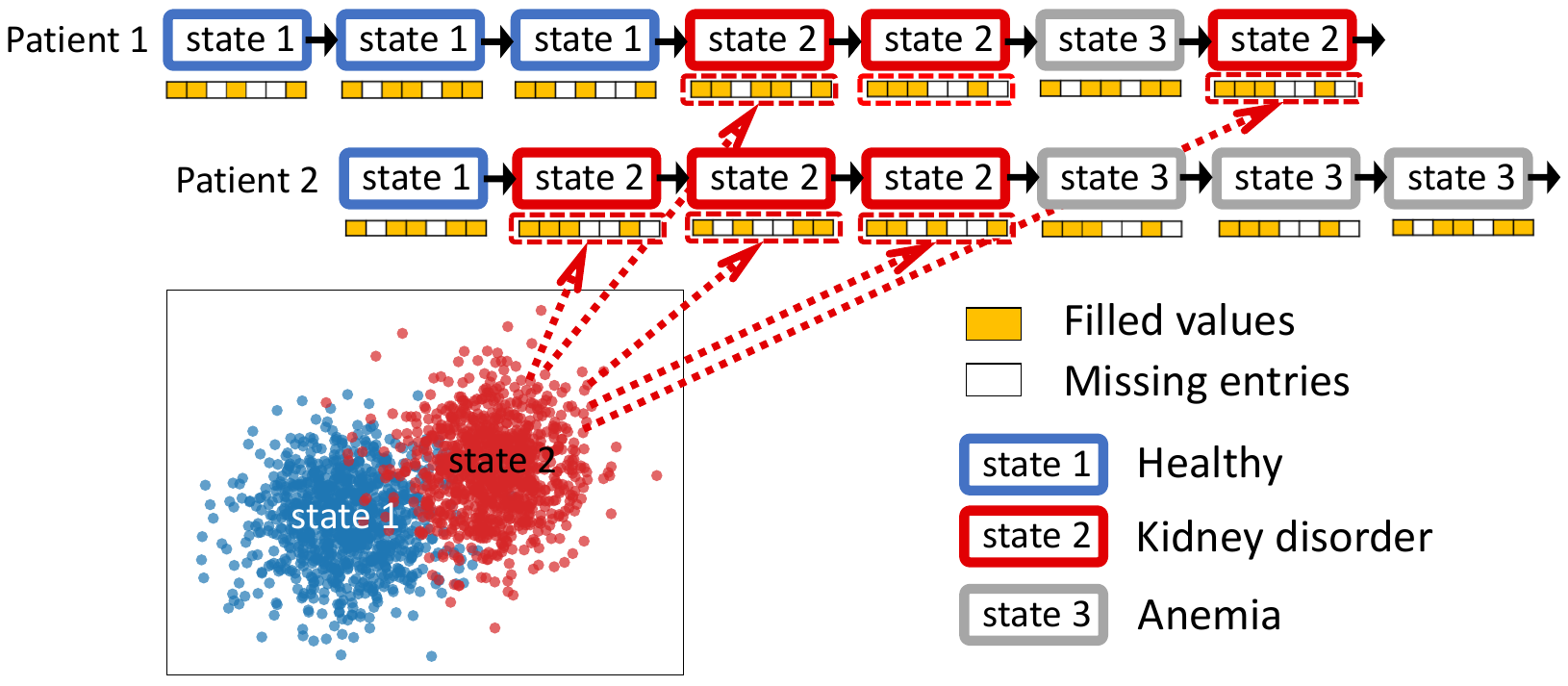}
\caption{An illustration of latent structures underlying the sparse MTS of two dialysis patients. The vector below each state is a temporal feature generated from some distribution.}\label{fig.intro}
\vspace{-0.35cm}
\end{figure}

For example, consider MTS signals collected for dialysis patients. Dialysis is an important renal replacement therapy 
for purifying the blood of patients whose kidneys are not working normally \cite{inaguma2019prediction}. Dialysis patients  have routines of dialysis, blood tests, chest X-ray, etc., 
which record data such as venous pressure, glucose level, and cardiothoracic ratio (CTR). These 
signal sources may have 
different frequencies. For instance, 
blood tests and CTR are often evaluated less frequently than dialysis. Different sources may not be aligned in time and what makes things worse is that some sources may be irregularly sampled, 
and missing entries may present. 
Despite such discrepancies, different signals give complementary views on a patient's physical condition, and therefore are all important to the diagnostic analysis.
However, simply combining the signals will induce highly sparse MTS data.
Similar scenarios are also found in other domains: In finance, time series from financial news, stock markets, and investment banks are collected at asynchronous time steps, but are correlated strongly \cite{binkowski2018autoregressive}. In large-scale complex 
monitoring systems, sensors of multiple sub-components may have different running environments, thus continuously emitting asynchronous time series that may still be related \cite{safari2014multirate}.

The sparsity of MTS signals when integrated from heterogeneous sources presents several challenges.
In particular, it complicates temporal dependencies and prevents popular models, such as recurrent neural networks (RNNs), from being directly used. The most common way to handle sparsity is to first impute missing values, and then make predictions on the imputed MTS. However, as discussed in \cite{che2018recurrent}, this two-step approach fails to account for the relationship between missing patterns and predictive tasks, 
leading to sub-optimal results when the sparsity is severe. 

Recently, some end-to-end models have been proposed. One approach is to consider missing time steps as intervals, and design RNNs with continuous dynamics via functional decays between observed time steps \cite{cao2018brits,rubanova2019latent}. 
Another approach is to parameterize all missed entries and jointly train the parameters with predictive models, so that the missing patterns are learned to fit downstream tasks \cite{che2018recurrent,shukla2019interpolation,tang2020joint}. However, these methods have the drawback that MTS samples are assessed individually. Latent relational structures that are shared by different MTS samples are seldom explored for robust modeling.

In many applications, MTS's are not independent, but are related by hidden 
structures. Fig. \ref{fig.intro} shows an example of two dialysis patients. Throughout the course of treatments, each patient may experience different latent states, such as kidney disorder and anemia, which are externalized by 
time series, 
such as glucose, albumin, and platelet levels. If two patients have similar pathological conditions, some of their data may be generated from similar state patterns, and could form clustering structures. 
Thus, inferring latent states and
modeling their dynamics 
are promising for leveraging the complementary information in clusters, which can 
alleviate the issue of sparsity. This idea is not limited to the medical domain. For example, in meteorology, nearby observing stations that monitor climate  may experience similar weather conditions ({\em i.e.}, latent states), 
which govern the generation of metrics, such as temperature and precipitation, over time. Although promising, inferring the latent clustering structures while modeling the dynamics underlying sparse MTS data is a challenging problem.

To address this problem,  we propose a novel 
\underline{D}ynamic \underline{G}aussian \underline{M}ixture based \underline{D}eep \underline{G}enerative \underline{M}odel (\ourmethod). 
\ourmethod\ has a state space model under a non-linear transition-emission framework. For each MTS, it models the transition of latent cluster variables, instead of isolated feature representations, 
where all transition distributions are parameterized by neural networks.  
\ourmethod\ is characterized by its emission step, where a dynamic Gaussian mixture distribution is proposed to capture the dynamics of clustering structures. For inductive analysis, we resort to variational inferences, and develop structured inference networks 
to approximate posterior distributions. To ensure reliable inferences, we also adopt the paradigm of parametric pre-imputation, 
and link a pre-imputation layer ahead of the inference networks. The  \ourmethod\ model is carefully designed to handle discrete variables and is constructed to be end-to-end trainable. 
Our contributions are summarized as follows:
\begin{itemize}
\item We investigate the problem of sparse MTS forecasting by modeling the latent dynamic clustering structures.
\item We propose \ourmethod, 
a novel deep generative model that leverages the transition of latent clusters and the emission from dynamic Gaussian mixture for robust forecasting.
\item We perform extensive experiments on real-life datasets to validate the 
effectiveness of our proposed method.
\end{itemize}

\section{Related Work}

To the best of our knowledge, this is the first work to exploit latent clustering structures via dynamic Gaussian mixture distributions for robust forecasting on sparse MTS.

Traditional forecasting methods are mainly developed for homogeneously complete MTS data, such as autoregression, ARIMA, and boosting trees \cite{chen2016xgboost}. Recently, to tackle non-linear temporal dynamics, various deep learning models have been proposed \cite{xingjian2015convolutional,qin2017dual}. These methods, however, are not designed to handle the challenges of highly sparse MTS. Their applicability relies heavily on pre-processing steps such as statistical imputation (e.g., mean imputation) \cite{che2018recurrent}, kernel based methods \cite{rehfeld2011comparison}, matrix completion \cite{koren2009matrix}, multivariate imputation by chained equations \cite{azur2011multiple}, and recent GAN based methods \cite{luo2018multivariate}. Such a two-step approach neglects the sparsity patterns that could be aligned with the downstream tasks, thus often leading to sub-optimal solutions on highly sparse MTS \cite{che2018recurrent}.


A more reasonable way is to apply end-to-end training methods on sparse MTS, 
which can be divided into two categories. The first is to transform the sparsity to time gaps between observations, which are integrated into the predictive models via (1) being explicit parts of the input features \cite{lipton2016modeling,binkowski2018autoregressive}, and (2) decaying the hidden states by exponential functions \cite{baytas2017patient,che2018recurrent}, or solving ordinary differential equations (ODEs) \cite{rubanova2019latent,de2019gru}. 
The second category trains a joint model for concurrent imputation and forecasting, so that task-aware missing patterns can be learned from the back-propagated errors.
For example, \citet{che2018recurrent} and \citet{tang2020joint} used a trainable decay mechanism for approximating missing values, \citet{cao2018brits} regarded unobserved entries as variables of a bidirectional RNN graph, \citet{shukla2019interpolation} exerted several kernel-based intensity functions to parameterize missing variables. In addition to these methods, \citet{che2018hierarchical} also studied a similar problem of modeling multi-rate MTS. However, none of the above methods explores the dynamic clustering structures underlying a batch of MTS samples for robust forecasting.

Vanilla Gaussian mixture (GM) model does not suit dynamic scenario. There are some works applying it on speech recognition \cite{tuske2015speaker,tuske2015integrating,variani2015gaussian,zhang2017joint}, which model the transition of words, but keep conditional GM distributions independent and static. \citet{diaz2018clustering} 
studied data streams of IoT systems, where a static GM model was continuously retrained to fit new data. In contrast to these methods, our model explicitly defines dynamic GM distributions with temporal dependencies, and is inductive and end-to-end trainable. Some dynamic topic models \cite{wei2007dynamic,zaheer2017latent} aim to unveil the flow of topics in documents. These methods, however, are neither Gaussian nor inductive, thus unable to be applied to solve the investigated problem.

\section{Problem Statement}

As suggested by the joint imputation-prediction framework \cite{che2018recurrent,shukla2019interpolation}, a sparse MTS sample can be represented with missing entries against a set of evenly spaced reference time points $t=1, ..., w$. 

Let $\bx_{1:w} = (\bx_1, ..., \bx_w) \in \mathbb{R}^{d \times w}$ be a length-$w$ MTS recorded from time steps $1$ to $w$, where $\bx_t=(x_{t}^{1}, ..., x_{t}^{d})^{\top} \in \mathbb{R}^{d}$ is a {\em temporal feature} vector at the $t$-th time step, $x_{t}^{i}$ is the $i$-th variable of $\bx_t$, and $d$ is the total number of variables. To mark observation times, we employ a binary mask $\bm_{1:w}=(\bm_1, \bm_2, ..., \bm_w) \in \{0, 1\}^{d \times w}$, where $m_{t}^{i}=1$ indicates $x_{t}^{i}$ is an observed entry; $m_{t}^{i}=0$ otherwise, with a corresponding placeholder $x_{t}^{i}=\texttt{NaN}$.

In this work, we are interested in a sparse MTS forecasting problem, which is to estimate the most likely length-$r$ sequence in the future given the {\em incomplete} observations in past $w$ time steps, 
{\em i.e.}, we aim to obtain
\begin{equation}
\mat{\tilde{x}}_{w+1:w+r} = \argmax_{\bx_{w+1:w+r}}p(\bx_{w+1:w+r}|\bx_{1:w}, \bm_{1:w})
\end{equation}
where $\mat{\tilde{x}}_{w+1:w+r}=(\mat{\tilde{x}}_{w+1}, ..., \mat{\tilde{x}}_{w+r})$ are predicted estimates, and $p(\cdot|\cdot)$ is a forecasting function to be learned.


\section{Our Proposed Model}

\begin{figure*}[!t]
\begin{center}
\includegraphics[width=1.95\columnwidth]{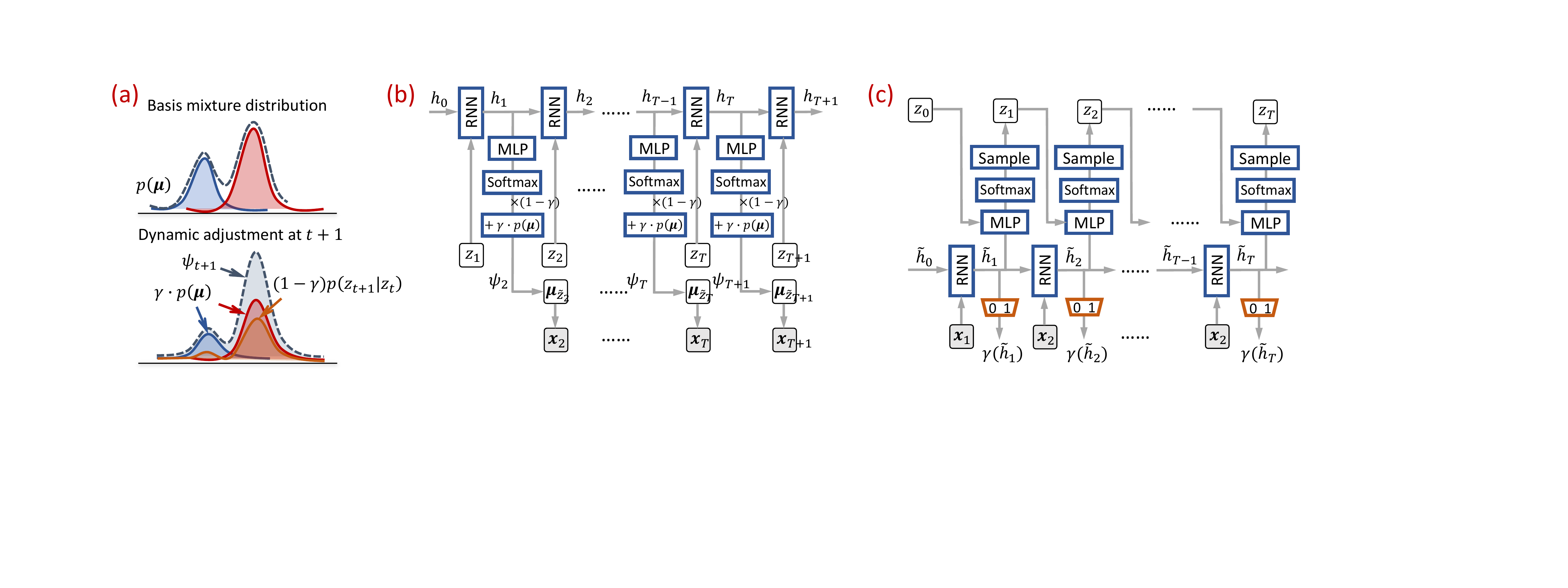}
\end{center}
\caption{Illustrative examples of (a) the dynamic adjustment of the Gaussian mixture according to Eq.~\eqref{eq.dynamic_probability}, with two mixture components, (b) the generative network, and (c) the inference network, where $\gamma(\cdot)$ is a gate function used in Eq.~\eqref{eq.dynamic_probability}.}\label{fig.method}
\vspace{-0.3cm}
\end{figure*}

In this section, we introduce our \ourmethod\ model. Inspired by the successful paradigm of joint imputation and prediction, 
we design \ourmethod\ to have a 
{\em pre-imputation layer} for capturing (1) the temporal intensity, and (2) the multi-dimensional correlations present in every MTS, for parameterizing missing entries. The parameterized MTS is fed to a {\em forecasting component}, which has a deep generative model that estimates the latent dynamic 
distributions for robust forecasting.


\subsection{Pre-Imputation Layer}


This layer 
aims to estimate the missing entries by leveraging the smooth trends 
and temporal intensities of the observed parts, 
which can help alleviate the impacts of sparsity in the downstream predictive tasks.

Similar to \cite{shukla2019interpolation}, for the $i$-th variable at the $t^{*}$-th reference 
time point, we use a Gaussian kernel $\kappa(t^{*}, t; \alpha_{i}) = e^{-\alpha_{i}(t^{*} - t)^{2}}$ to evaluate the temporal influence of any time step $t$ ($1\le t\le w$) on $t^{*}$, where $\alpha_{i}$ is a parameter to be learned. Based on the kernel, we then employ a weighted aggregation for estimating $x_{t^{*}}^{i}$ by
\begin{equation}\label{eq.alpha}
\bar{x}_{t^{*}}^{i} = \frac{1}{\lambda(t^{*}, \mat{m}^{i}; \alpha_{i})}\sum_{t=1}^{w}\kappa(t^{*}, t; \alpha_{i})m_{t}^{i}x_{t}^{i}
\end{equation}
where $\mat{m}^{i}=(m_{1}^{i}, ..., m_{w}^{i})^{\top}\in\mathbb{R}^{w}$ is the mask of the $i$-th variable, and $\lambda(t^{*}, \mat{m}^{i}; \alpha_{i}) = \sum_{t=1}^{w}m_{t}^{i}\kappa(t^{*}, t; \alpha_{i})$ is an intensity function that evaluates the observation density at $t^{*}$, in which $m_{t}^{i}$ is used to zero out unobserved time steps.

To account for the correlations of different variables, we also merge the information across $d$ variables by introducing learnable correlation coefficients $\rho_{ij}$ for $i,j=1, ..., d$, and formulating a parameterized output if ${x}_{t^{*}}^{i}$ is unobserved.
\begin{equation}\label{eq.rho}
\small
\hat{x}_{t^{*}}^{i} = \bigg[\sum_{j=1}^{d}\rho_{ij}\lambda(t^{*}, \mat{m}^{i}; \alpha_{j})\bar{x}_{t^{*}}^{j}\bigg]/\sum_{j'=1}^d\lambda(t^{*}, \mat{m}^{i}; \alpha_{j'})
\end{equation}
where 
$\rho_{ij}$ is set as $1$ for $i=j$, and $\lambda(t^{*}, \mat{m}^{i}; \alpha_{j})$ is introduced to indicate the reliability of $\bar{x}_{t^{*}}^{j}$, because larger $\lambda(t^{*}, \mat{m}^{i}; \alpha_{j})$ implies more observations near $\bar{x}_{t^{*}}^{j}$. 

In this layer, the set of parameters are $\boldsymbol{\alpha}=[\alpha_{1}, ..., \alpha_{d}]$, and $\boldsymbol{\rho}=[\rho_{ij}]_{i,j=1}^{d}\in\mathbb{R}^{d \times d}$. \ourmethod\ trains them jointly with its generative model for aligning missing patterns with the forecasting tasks.

\subsection{Forecasting Component}

Next, we design a generative model that captures the latent dynamic clustering structures for robust forecasting.

Suppose there are $k$ latent clusters underlying all temporal features $\mat{x}_{t}$'s in a batch of MTS samples. For every time step $t$, we associate $\mat{x}_{t}$ with a {\em latent cluster variable} $z_{t}$ to indicate to which cluster $\mat{x}_{t}$ belongs. Instead of the transition of 
$\mat{x}_{t}\rightarrow\mat{x}_{t+1}$, in this work, we propose to model the transition of the cluster variables $z_{t}\rightarrow\z_{t+1}$. 
Since the clusters integrate the complementary information of similar features across MTS samples at different time steps, leveraging them is more robust than using individual sparse feature $\mat{x}_{t}$'s.

\subsubsection{Generative Model.}
The generative process of our \ourmethod\ follows the {\em transition} and {\em emission} framework of state space models \cite{krishnan2016structured}.

First, the transition process of \ourmethod\ 
employs a recurrent structure 
due to its effectiveness on modeling long-term temporal dependencies of sequential variables. Each time, the probability of a new state $z_{t+1}$ is updated upon its previous states $z_{1:t}=(z_{1}, ..., z_{t})$. We use a learnable function to define the transition probability, {\em i.e.}, 
$p(z_{t+1}|z_{1:t})=f_{\boldsymbol{\theta}}(z_{1:t})$, 
where the function $f_{\boldsymbol{\theta}}(\cdot)$ is parameterized by $\boldsymbol{\theta}$, which can be variants of RNNs, for encoding non-linear dynamics that may be established between the latent variables.

For the emission process, we propose a {\em dynamic Gaussian mixture} distribution, which is defined by dynamically tuning a static basis mixture distribution. Let $\boldsymbol{\mu}_{i}$ ($i=1, ..., k$) be the mean of the $i$-th mixture component of the basis distribution, and 
$p(\boldsymbol{\mu}_{i})$ be its corresponding mixture probability. The emission (or forecasting) of a new feature $\mat{x}_{t+1}$ at time step $t+1$ involves two steps: (1) draw a latent cluster variable $z_{t+1}$ from a categorical distribution on all mixture components, and (2) draw $\mat{x}_{t+1}$ from the Gaussian distribution $\mathcal{N}(\boldsymbol{\mu}_{\z_{t+1}}, \sigma^{-1}\mat{I})$, where $\sigma$ is a hyperparameter, and $\mat{I}$ is an identity matrix. Here, we use isotropic Gaussian because of its efficiency and effectiveness in our experiments.

In step (1), the categorical distribution is usually defined on $p(\boldsymbol{\mu})=[p(\boldsymbol{\mu}_{1}), ..., p(\boldsymbol{\mu}_{k})]\in\mathbb{R}^{k}$, {\em i.e.}, the static mixture probabilities, which cannot reflect the dynamics in MTS. In light of this, and considering the fact that transition probability $p(z_{t+1}|z_{1:t})$ indicates to which cluster $\mat{x}_{t+1}$ belongs, we propose to dynamically adjust the mixture probability at each time step using $p(z_{t+1}|z_{1:t})$ by
\begin{equation}\label{eq.dynamic_probability}
    \boldsymbol{\psi}_{t+1}=\underbrace{(1-\gamma)p(z_{t+1}|z_{1:t})}_{\text{dynamic adjustment}} + \underbrace{\gamma p(\boldsymbol{\mu})}_{\text{basis mixture}}
\end{equation}
where $\boldsymbol{\psi}_{t+1}$ 
is the dynamic mixture distribution at time step $t+1$, and $\gamma$ is a hyperparameter within $[0, 1]$ that controls the relative degree of change that deviates from the basis mixture distribution. 

Fig. \ref{fig.method}(a) illustrates 
the dynamic adjustment process of Eq.~\eqref{eq.dynamic_probability} 
on a Gaussian mixture with two components, where $p(z_{t+1}|z_{1:t})$ adjusts the mixture towards the component ({\em i.e.}, cluster) that $\mat{x}_{t+1}$ belongs to. It is noteworthy that adding the basis mixture in Eq.~\eqref{eq.dynamic_probability} is indispensable because it determines the relationships between different components, which regularizes the learning of the means $\boldsymbol{\mu}=[\boldsymbol{\mu}_{1}, ..., \boldsymbol{\mu}_{k}]$ during model training.

As such, our generative process can be summarized as
\begin{outline}[enumerate]
\small
\1 for each MTS sample:
    \2 draw $z_{1} \sim \text{Uniform}(k)$
    \2 for time step $t=1, ..., w$:
        \3 compute transition probability by $p(z_{t+1}|z_{1:t})=f_{\boldsymbol{\theta}}(z_{1:t})$
        \3 draw $z_{t+1}\sim\text{Categorial}(p(z_{t+1}|z_{1:t}))$ for transition
        \3 draw 
        $\tilde{z}_{t+1}\sim\text{Categorial}(\boldsymbol{\psi}_{t+1})$ 
        using Eq.~\eqref{eq.dynamic_probability} for emission
        \3 draw a feature vector $\mat{\tilde{x}}_{t+1}\sim\mathcal{N}(\boldsymbol{\mu}_{\tilde{z}_{t+1}}, \sigma^{-1}\mathbf{I})$
\end{outline}
where $z_{t+1}$ (step ii) and $\tilde{z}_{t+1}$ (step iii) are different: $z_{t+1}$ is used in transition (step i) for maintaining recurrent property; 
$\tilde{z}_{t+1}$ is used in emission from updated mixture distribution.

In the above process, the parameters in $\boldsymbol{\mu}_{i}$ are shared by samples in the same cluster, whereby consolidating complementary information for robust forecasting.

\subsubsection{Parameterization of Generative Model.}
The key parametric function in the generative process is $f_{\boldsymbol{\theta}}(\cdot)$, for which we choose a recurrent neural network architecture as
\begin{equation}\label{eq.gen_rnn}
\begin{aligned}
&p(z_{t+1}|z_{1:t})=\text{softmax}(\text{MLP}(\mat{h}_{t})),\\
&\text{where}~~\mat{h}_{t}=\text{RNN}(z_{t}, \mat{h}_{t-1}),
\end{aligned}
\end{equation}
and $\mat{h}_{t}$ is the $t$-th hidden state, MLP represents a multilayer perceptron, RNN can be instantiated by either an LSTM or a GRU. Moreover, to accommodate the applications where the reference time steps of MTS's could be unevenly spaced, we can also incorporate the recently proposed neural ordinary differential equations (ODE) based RNNs \cite{rubanova2019latent} to handle the time intervals. In our experiments, we demonstrate the flexibility of our framework in Eq.~\eqref{eq.gen_rnn} by evaluating several choices of RNNs.

Fig. \ref{fig.method}(b) illustrates our generative network. 
In summary, the set of trainable parameters of the generative model is $\boldsymbol{\vartheta}=\{\boldsymbol{\theta}, \boldsymbol{\mu}\}$. Given this, we aim at maximizing the log marginal likelihood of observing each MTS sample, {\em i.e.},
\begin{equation}\label{eq.marginal}
\begin{aligned}
&\mathcal{L}(\boldsymbol{\vartheta})=\log(\sum_{z_{1:w}}p_{\boldsymbol{\vartheta}}(\mat{x}_{1:w}, z_{1:w}))
\end{aligned}
\end{equation}
where the joint probability in Eq.~\eqref{eq.marginal} can be factorized w.r.t. the dynamic mixture distribution in Eq.~\eqref{eq.dynamic_probability} after the Jensen's inequality is applied on $\mathcal{L}(\boldsymbol{\vartheta})$ by
\begin{equation}\label{eq.marginal_factor}
\small
\begin{aligned}
\mathcal{L}(\boldsymbol{\vartheta})\geq\sum_{t=0}^{w-1}\sum_{z_{1:t+1}}&\bigg[\log\big(p_{\boldsymbol{\vartheta}}(\mat{x}_{t+1}|z_{t+1})\big)p_{\boldsymbol{\theta}}(z_{1:t}) \\
&\big[(1-\gamma)p_{\boldsymbol{\theta}}(z_{t+1}|z_{1:t}) + \gamma p(\boldsymbol{\mu}_{z_{t+1}})\big]\bigg]
\end{aligned}
\end{equation}
in which the above lower bound will serve as our objective to be maximized. The detailed derivation of Eq.~\eqref{eq.marginal_factor} is deferred to the supplementary materials.

In order to estimate the parameters $\boldsymbol{\vartheta}$, our goal is to maximize 
Eq.~\eqref{eq.marginal_factor}. However, summing out $z_{1:t+1}$ over all possible sequences is computationally difficult.
Therefore, evaluating the true posterior density $p(\mat{z}|\mat{x}_{1:w})$ is intractable. To circumvent this problem, meanwhile enabling inductive analysis, we resort to variational inference \cite{hoffman2013stochastic} and introduce an inference network in the following.

\subsubsection{Inference Network.}
We introduce an approximated posterior $q_{\boldsymbol{\phi}}(\mat{z}|\mat{x}_{1:w})$, which is parameterized by neural networks with parameters $\boldsymbol{\phi}$. 
We design our inference network to be structural, and 
employ the idea of deep Markov processes to maintain the temporal dependencies between latent variables, which leads to the following factorization.
\begin{equation}
\begin{aligned}
q_{\boldsymbol{\phi}}(\mat{z}|\mat{x}_{1:w})=q_{\boldsymbol{\phi}}(z_{1}|\mat{x}_{1})\prod_{t=1}^{w-1}q_{\boldsymbol{\phi}}(z_{t+1}|\mat{x}_{1:t+1}, z_{t})
\end{aligned}
\end{equation}


With the introduction of $q_{\boldsymbol{\phi}}(\mat{z}|\mat{x}_{1:w})$, instead of maximizing the log marginal likelihood $\mathcal{L}(\boldsymbol{\vartheta})$, we are interested in maximizing the variational {\em evidence lower bound} (ELBO) $\ell(\boldsymbol{\vartheta}, \boldsymbol{\phi}) \le \mathcal{L}(\boldsymbol{\vartheta})$ with respect to both $\boldsymbol{\vartheta}$ and $\boldsymbol{\phi}$. 
By incorporating the bounding step in Eq.~\eqref{eq.marginal_factor}, we can derive the EBLO of our problem, which is written by
\begin{equation}\label{eq.obj}
\footnotesize
\begin{aligned}
&\ell(\boldsymbol{\vartheta}, \boldsymbol{\phi}) = (1-\gamma)\sum_{t=1}^{w}\E_{q_{\boldsymbol{\phi}}(\z_t|\mat{x}_{1:t})}[\log{(p_{\boldsymbol{\vartheta}}(\bx_t|\z_t))}]\\
&-\sum_{t=1}^{w-1}\E_{q_{\boldsymbol{\phi}}(z_{1:t}|\mat{x}_{1:t})}[\mathcal{D}_{KL}\big(q_{\boldsymbol{\phi}}(z_{t+1}|\mat{x}_{1:t+1}, z_{t}) || p_{\boldsymbol{\vartheta}}(\z_{t+1}|\z_{1:t})\big)]\\
&-\mathcal{D}_{KL}\big(q_{\boldsymbol{\phi}}(\z_{1}|\mat{x}_{1}) || p_{\boldsymbol{\vartheta}}(z_1)\big) + \gamma\sum_{t=1}^{w}\sum_{z_{t}=1}^{k}p_{\boldsymbol{\vartheta}}(\boldsymbol{\mu}_{z_{t}})\log{(p_{\boldsymbol{\vartheta}}(\mat{x}_{t}|z_{t}))}
\end{aligned}
\end{equation}
where $\mathcal{D}_{KL}(\cdot||\cdot)$ is the KL-divergence. $p_{\boldsymbol{\vartheta}}(z_{1})$ is a uniform prior as described in the generative process. Similar to VAE \cite{kingma2013auto}, it helps prevent overfitting and improve the generalization capability of our model. The detailed derivation of Eq.~\eqref{eq.obj} can be found in the supplementary materials.

Eq.~\eqref{eq.obj} also sheds some insights on how our dynamic mixture distribution in Eq.~\eqref{eq.dynamic_probability} works: the first three terms encapsulate the learning criteria for dynamic adjustments; the last term after $\gamma$ regularizes the relationships between different basis mixture components.

Fig. \ref{fig.method}(c) illustrates the architecture of the inference network, in which $q_{\boldsymbol{\phi}}(z_{t+1}|\mat{x}_{1:t+1}, z_{t})$ is a recurrent structure
\begin{equation}\label{eq.infer_rnn}
\begin{aligned}
&q_{\boldsymbol{\phi}}(z_{t+1}|\mat{x}_{1:t+1}, z_{t})=\text{softmax}(\text{MLP}(\mat{\tilde{h}}_{t+1})),\\
&\text{where}~~\mat{\tilde{h}}_{t+1}=\text{RNN}(\mat{x}_{t}, \mat{\tilde{h}}_{t}),
\end{aligned}
\end{equation}
$\mat{\tilde{h}}_{t}$ is the $t$-th latent state of the RNNs, and $z_{0}$ is set to $\mat{0}$ so that it has no impact in the iteration.

Since sampling discrete variable $z_{t}$ from the categorical distributions in Fig. \ref{fig.method}(c) is not differentiable, it is difficult to optimize the model parameters. To get rid of it, we employ the Gumbel-softmax reparameterization trick \cite{jang2017categorical} to generate differetiable discrete samples, which is illustrated by the ``sample'' steps in Fig. \ref{fig.method}(c). In this way, our \ourmethod\ model is end-to-end trainable.

\subsubsection{Gated Dynamic Distributions.}
In Eq.~\eqref{eq.dynamic_probability}, the dynamics of the Gaussian mixture distribution is tuned by a hyperparameter $\gamma$, which may require some tuning efforts on validation datasets. To circumvent it, we introduce a gate function $\gamma(\mat{\tilde{h}}_{t})=\text{sigmoid}(\text{MLP}(\mat{\tilde{h}}_{t}))$ using the information extracted by the inference network, as shown in Fig. \ref{fig.method}(c), to substitute $\gamma$ in Eq.~\eqref{eq.dynamic_probability}. As such, $\boldsymbol{\psi}_{t}$ becomes a gated distribution that can be dynamically tuned at each time step.

\subsection{Model Training}

We jointly learn the parameters $\{\boldsymbol{\alpha}, \boldsymbol{\rho}, \boldsymbol{\vartheta}, \boldsymbol{\phi}\}$ of the pre-imputation layer, the generative network $p_{\boldsymbol{\vartheta}}$, and the inference network $q_{\boldsymbol{\phi}}$ by maximizing the ELBO in Eq.~\eqref{eq.obj}.

The main challenge to evaluate Eq.~\eqref{eq.obj} is to obtain the gradients of all terms under the expectation $\mathbb{E}_{q_{\boldsymbol{\phi}}}$. Because $z_{t}$ is categorical, the first term can be analytically calculated with the probability $q_{\boldsymbol{\phi}}(z_{t}|\mat{x}_{1:t})$. However, $q_{\boldsymbol{\phi}}(z_{t}|\mat{x}_{1:t})$ is not an output of the inference network, so we derive a subroutine to compute $q_{\boldsymbol{\phi}}(z_{t}|\mat{x}_{1:t})$ from $q_{\boldsymbol{\phi}}(z_{t}|\mat{x}_{1:t}, z_{t-1})$. In the second term, since the KL divergence is sequentially evaluated, we employ ancestral sampling techniques to sample $z_{t}$ from time step $1$ to $w$ to approximate the distribution $q_{\boldsymbol{\phi}}$. 
It is also noteworthy that in Eq.~\eqref{eq.obj}, we only evaluate observed values in $\mat{x}_{t}$ by using masks $\mat{m}_{t}$ to mask out the unobserved parts. The subroutine to compute $q_{\boldsymbol{\phi}}(z_{t}|\mat{x}_{1:t})$ can be found in the supplementary materials.

As such, the entire \ourmethod\ is differentiable, and we use stochastic gradient descents to optimize Eq.~\eqref{eq.obj}. In the last term of Eq.~\eqref{eq.obj}, we also need to perform density estimation of the basis mixture distribution, {\em i.e.}, to estimate $p(\boldsymbol{\mu})$. Given a batch of MTS samples, suppose there are $n$ temporal features $\mat{x}_{t}$ in this batch, and their collection is denoted by a set $\mathcal{X}$, we can estimate the mixture probability by
\begin{equation}
\small
\begin{aligned}
p(\boldsymbol{\mu}_{i})=\sum_{\mat{x}_{t}\in\mathcal{X}}q_{\boldsymbol{\phi}}(z_{t}=i|\mat{x}_{1:t}, z_{t-1})/n,~~\text{for}~i=1, ..., k
\end{aligned}
\end{equation}
where $q_{\boldsymbol{\phi}}(z_{t}=i|\mat{x}_{1:t}, z_{t-1})$ is the inferred membership probability of $\mat{x}_{t}$ to the $i$-th latent cluster by Eq.~\eqref{eq.infer_rnn}.

\section{Experiments}

\begin{table*}
\centering
\small
\caption{Forecasting results (RMSE and MAE) of the compared methods on different datasets}\label{Table: forecasting_res}
\vspace{-0.2cm}
\begin{tabular}[!h]{>{\arraybackslash}p{1.4cm}|>{\centering\arraybackslash}p{2.1cm}>{\centering\arraybackslash}p{2.1cm}|>{\centering\arraybackslash}p{2.1cm}>{\centering\arraybackslash}p{2.1cm}|>{\centering\arraybackslash}p{2.1cm}>{\centering\arraybackslash}p{2.1cm}} \hline
\multirow{2}{*}{Method} &\multicolumn{2}{c|}{\mimic}&\multicolumn{2}{c|}{\climate}&\multicolumn{2}{c}{KDD-CUP} \\ \hhline{~------}
& RMSE & MAE& RMSE & MAE& RMSE & MAE\\ \hline
\var & 6.7154$\pm$0.0360 & 3.7420$\pm$0.0246 & 0.9811$\pm$0.0183 & 0.7927$\pm$0.0195 & 0.8164$\pm$0.0377 & 0.5052$\pm$0.0256\\
\lstm & 1.0587$\pm$0.0091 & 0.8203$\pm$0.0106 & 0.6340$\pm$0.0098  & 0.4456$\pm$0.0098 & 0.7465$\pm$0.0370 & 0.5082$\pm$0.0358\\
\dmm & 0.9852$\pm$0.0025 & 0.7510$\pm$0.0057 & 0.6068$\pm$0.0092 & 0.4058$\pm$0.0112 & 0.7067$\pm$0.0220 & 0.5052$\pm$0.0303\\
\xgboost & 0.9900$\pm$0.0002 &0.7209$\pm$0.0003  & 0.8664$\pm$0.0011 & 0.7816$\pm$0.0010  & 0.8841$\pm$0.0106 & 0.7580$\pm$0.0102\\
\grui & 1.0322$\pm$0.0069 & 0.8256$\pm$0.0074 & 0.9491$\pm$0.0046 & 0.7764$\pm$0.0049 & 0.8233$\pm$0.0333 & 0.5925$\pm$0.0364\\
\grud & 1.0495$\pm$0.0068 & 0.8502$\pm$0.0069 & 0.9695$\pm$0.0089 & 0.7912$\pm$0.0087 & 0.7268$\pm$0.0254 & 0.5051$\pm$0.0228\\
IPN & 0.9888$\pm$0.0025 & 0.7856$\pm$0.0039 & 0.6097$\pm$0.0066 & 0.4204$\pm$0.0089 & 0.7207$\pm$0.0313 & 0.4834$\pm$0.0229\\
\lgnet & 0.9590$\pm$0.0033 & 0.7093$\pm$0.0033 & 0.5883$\pm$0.0071 & \underline{0.3841$\pm$0.0063} & 0.7346$\pm$0.0272 & 0.5181$\pm$0.0218\\ 
\lode & 0.9315$\pm$0.0034 & 0.7325$\pm$0.0035 & 0.6171$\pm$0.0056 & 0.4216$\pm$0.0038 & 0.8226$\pm$0.0387 & 0.5834$\pm$0.0331\\\hline
\ourmethod-L & \underline{0.9143$\pm$0.0025} & \underline{0.7089$\pm$0.0037} & \underline{0.5426$\pm$0.0066} & 0.3848$\pm$0.0047 & \underline{0.6975$\pm$0.0224} & \underline{0.4748$\pm$0.0162}\\
\ourmethod-O & \textbf{0.9003$\pm$0.0015} & \textbf{0.6876$\pm$0.0027} & \textbf{0.4983$\pm$0.0053} & \textbf{0.3367$\pm$0.0059} & \textbf{0.6835$\pm$0.0276} & \textbf{0.4646$\pm$0.0213}\\ \hline
\end{tabular}
\end{table*}

\begin{table}[!t]
\centering
\small
\caption{Imputation results (RMSE) before/after the forecasting component of \ourmethod-L (-L) and \ourmethod-O (-O)}\label{tab.imputation}
\vspace*{-0.2cm}
\begin{tabular}{ll|c|c|c} \hline
\multicolumn{2}{l|}{Setting} &\mimic & \climate & KDD-CUP\\ \hline
\multirow{2}{*}{-L} & before & 1.4111 & 0.7780 & 5.2363\\
& after & \textbf{0.9052} & \textbf{0.5250} & \textbf{0.5506} \\ \hline
\multirow{2}{*}{-O} & before &1.4186 & 0.4761& 4.2868 \\
& after & \textbf{0.8979} & \textbf{0.4663} & \textbf{0.5362} \\ \hline
\end{tabular}
\end{table}

\begin{table}[!t]
\centering
\small
\caption{Ablation analysis (RMSE)}\label{Table: gate_forecasting_res_ode}
\vspace*{-0.2cm}
\begin{tabular}{l|c|c|c} \hline
Model &\mimic & \climate & KDD-CUP\\ \hline
(a) $\gamma=1.0$ & 0.9832 &0.9913& 0.9998 \\
(b) $\gamma=0.0$ & 0.9191 & 0.5151 & 0.7533\\
(c) $\gamma=1e^{-2}$ & \underline{0.9033} & \textbf{0.4958} & \underline{0.6878}\\
(d) Gate $\gamma(\cdot)$ & \textbf{0.9003} & \underline{0.4983} & \textbf{0.6835}\\ \hline
\end{tabular}
\end{table}

In this section, we evaluate the performance of our \ourmethod\ model on real-life datasets from different domains and compare it with state-of-the-art approaches.

\subsection{Datasets and Task Description}

\subsubsection{USHCN\footnote{https://www.ncdc.noaa.gov/ushcn/introduction}}
This dataset consists of daily meteorological records collected from 1219 stations across the contiguous states from 1887 to 2009. It has five climate features, such as mean temperature and total daily precipitation. From the dataset, we randomly extracted 5000 segments recorded at all stations in NY state from 1900 to 2000. Each segment is an MTS sample with 100 consecutive daily records. The dataset has a missing ratio of 10.4\%. Our task is to use the past 80 days' records to forecast the future 20 days' weather.

\subsubsection{KDD-CUP\footnote{https://www.kdd.org/kdd2018/kdd-cup}}
This is an air quality dataset from KDD CUP challenge 2018, which consists of PM2.5 measurements from 35 monitoring stations in Beijing. The values were recorded hourly from 01/2017 to 12/2017, 
and the overall missing ratio is 16.5\%. On this dataset, our task is to forecast the PM2.5 values for the 35 stations in future 12 hours using the historical records in the past 24 hours.

\subsubsection{MIMIC-III}
This is a public clinical dataset \cite{johnson2016mimic}, with over 58,000 hospital admission records. We collected 31,332 adults' ICU stay data, 
and extracted 17 clinical features, such as glucose and heart rate, from the first 72 hours using the benchmark tool \cite{Harutyunyan2019}. 
Because of the heterogeneous sources (e.g., lab tests, routine signals, etc.), the MTS's are highly sparse, with a missing ratio of 72.7\%. The task is to forecast the last 24 hour signals using the recorded data in the first 48 hours.

\subsection{Compared Methods}
We compare our model with both conventional approaches and state-of-the-art approaches, including Vector Autoregression (VAR), LSTM, Deep Markov Model (DMM) \cite{krishnan2016structured}, \xgboost\ \cite{chen2016xgboost}, \grui\ \cite{luo2018multivariate}, GRU-D \cite{che2018recurrent}, Interpolation-Prediction Networks (IPN) \cite{shukla2019interpolation}, LGNet \cite{tang2020joint}, and Latent ODE (L-ODE) \cite{rubanova2019latent}.

Among these methods, VAR, \xgboost, LSTM and DMM cannot handle missing values. Therefore, we follow \cite{tang2020joint} to concatenate each MTS with its mask matrix as their input features. \grui\ is a two-step approach, which first imputes missing values with GAN-based networks, and then uses GRU for forecasting. GRU-D, IPN, and LGNet deal with sparsity by jointly training parametic imputation functions and RNNs. GRU-D and LGNet use exponential decay based imputation, IPN uses kernel intensity based functions. In addition, 
LGNet has a memory module to leverage global MTS patterns in its RNNs. Different from the above methods, L-ODE addresses sparsity by modeling uneven time intervals via learnable ordinary differential equations. If a temporal feature $\mat{x}_{t}$ is too sparse ({\em e.g.}, the sparsity is above a threshold), L-ODE will replace it by an interval, thus generating unevenly spaced MTS's. It is worth noting that none of these methods exploits the clustering structures underlying the MTS set.

For our method, since it is a flexible framework, we use LSTM and ODE as the RNNs, and denote them by \ourmethod-L and \ourmethod-O, where the latter has the capability to handle uneven time intervals. To gain further insights on some of the design choices, we also compare \ourmethod\ with its variants, which will be discussed in the ablation analysis.

\subsection{Experimental Setup}

For each dataset, train/valid/test sets were split as 70/10/20. 
All compared methods were trained by Adam optimizer with 
hyperparameters selected on the validation set. The configurations of the compared methods are described in the supplementary materials. For \ourmethod, we grid-searched $k$, {\em i.e.}, the number of mixture components (or $\boldsymbol{\mu}_{i}$'s), from 10 to 200. The $\gamma$ in Eq.~\eqref{eq.dynamic_probability} was searched within \{$1e^{-5}$, $1e^{-4}$, $1e^{-3}$, $1e^{-2}$, $1e^{-1}$\}. The gate function $\gamma(\cdot)$ was also tested to automatically tune Eq.~\eqref{eq.dynamic_probability}. The variance $\sigma$ of Gaussian distributions was selected from \{$1e^{-5}$, $1e^{-4}$, $1e^{-3}$, $1e^{-2}$, $1e^{-1}$\}. Similar to \cite{krishnan2016structured}, we configured \ourmethod-L with one-layer LSTMs, and searched the hidden dimensionality within $\{10, 20, 30, 40, 50\}$.
For \ourmethod-O, we follow \cite{rubanova2019latent} to use an MLP for instantiating the neural networks, with dimensionality selected from $\{10, 20, 30, 40, 50\}$. The parameters in $\boldsymbol{\mu}$ were randomly initialized. Early stopping was applied to avoid overfitting.


To evaluate the performance of the compared methods, we use the widely used root mean square error (\rmse) and mean absolute error (\mae) \cite{che2018hierarchical}. For both metrics, smaller values indicate better performance.


\subsection{Experimental Results}

\begin{figure}[!t]
\begin{center}
\centerline{\includegraphics[width=1.0\columnwidth, height=0.35\columnwidth]{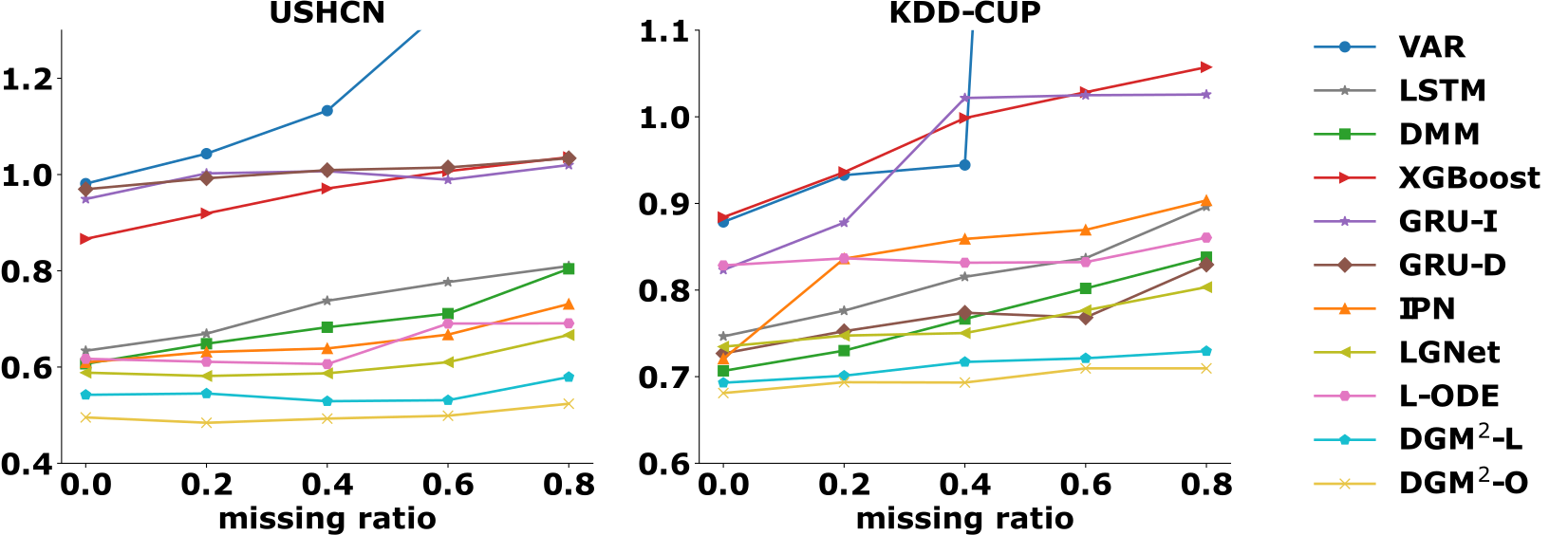}}
\caption{Forecasting results (\rmse) of the compared methods w.r.t. varying missing ratios on \climate\ and \beijing
}
\label{fig: beijing_forecast_robust}
\end{center}
\vspace{-0.3cm}
\end{figure}

\begin{figure*}[!t]
\begin{center}
\includegraphics[width=1.9\columnwidth]{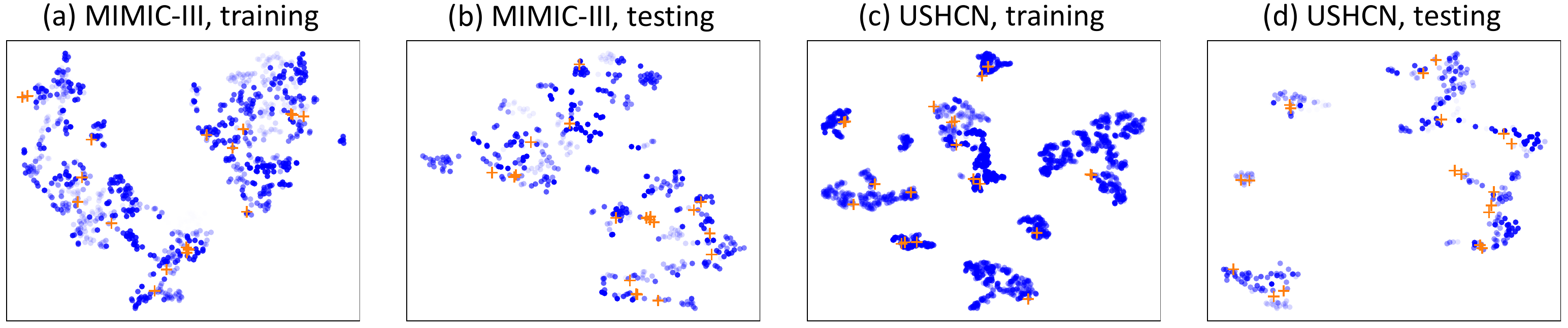}
\end{center}
\caption{The tSNE visualization on MIMIC-III and USHCN datasets. Circles are temporal features. Different transparencies indicate different MTS samples from which the features are extracted. ``+'' markers are the learned Gaussian means by \ourmethod.}\label{fig.vis}
\vspace{-0.3cm}
\end{figure*}

\subsubsection{Forecasting}

Table \ref{Table: forecasting_res} summarizes the average results of the compared methods over 5 runs, from which we have several observations. First, methods under joint imputation-prediction framework, such as IPN and LGNet, often outperform traditional methods. This validates the usefulness of learning task-aware missing patterns. Second, DMM performs well even without special designs for handling sparsity, which suggests the superiority of generative models on the forecasting tasks. We also observe good results of L-ODE on USHCN and MIMIC-III. This may indicate the effectiveness of modeling time intervals as another way toward handling sparse MTS's. Finally, the proposed \ourmethod-L outperforms other methods in most cases, which validates its design as a generative model with a joint imputation-prediction paradigm. More importantly, this justifies its dynamic modeling of latent clustering structures. Furthermore, our flexible framework enables \ourmethod-O, which incorporates the advantages of ODE. It obtains further improvements, with relative improvements on RMSE of at least 3.5\%, 18.1\% and 3.4\% on MIMIC-III, USHCN, and KDD-CUP datasets, respectively. The statistics in Table \ref{Table: forecasting_res} also demonstrate that the improvements are significant.

To gain further insights about the usefulness of modeling clustering structures, we compare the outputs of the pre-impulation layer and the forecasting component of \ourmethod. Since the forecasting component generates values at every time step, the generated values 
at those missing entries, {\em i.e.}, $x_{t}^{i}=\texttt{NaN}$, can be regarded as new imputations. If the newly imputed values are more accurate than the pre-imputed values, then it implies the learned clustering structures can facilitate generating true MTS's effectively. To this end, we randomly removed 10\% of the observations in each dataset, and evaluated the imputation error. Table \ref{tab.imputation} summarizes the average results of 5 runs. From Table \ref{tab.imputation}, we can observe both variants of \ourmethod\ significantly reduce RMSE via the generative model, which hence validates the usefulness of the learned clustering structures.

\subsubsection{Robustness}
Next, we evaluate the robustness of \ourmethod\ w.r.t. varying missing ratios using USHCN and KDD-CUP datasets, which have moderate sparsity. 
Specifically, we randomly dropped $\delta$ ($0\le\delta\le1$) fraction of the observed values, and tuned $\delta$ from 0 to 0.8. The compared methods were trained on the corrupted datasets, and evaluated on the same forecasting tasks as before. Fig. \ref{fig: beijing_forecast_robust} reports the results in terms of RMSE. First of all, Fig. \ref{fig: beijing_forecast_robust} shows that the forecasting errors of the methods without handling sparsity, such as VAR, LSTM, DMM, go up quickly as $\delta$ increases. In comparison, methods that explicitly address sparsity, {\em e.g.}, GRU-D, LGNet, and L-ODE, can maintain relatively stabler results to varying extents. However, many of them still suffer when the missing ratio is very high, {\em e.g.}, $\delta\ge 0.6$. In such scenarios, both \ourmethod-L and \ourmethod-O obtain the best forecasting accuracy, which is attributed to the modeling of robust clustering structures via dynamic Gaussian mixtures.

\subsubsection{Ablation Analysis}
In this section, we focus on the analysis of the gating mechanism $\gamma(\cdot)$ introduced in our inference network. Recall in Eq.~\eqref{eq.dynamic_probability}, $\gamma$ is a hyperparameter controlling the dynamics in the Gaussian mixture. Table \ref{Table: gate_forecasting_res_ode} compares the results of different choices on $\gamma$, as well as the use of $\gamma(\cdot)$, in terms of RMSE. $\gamma=0$ and $\gamma=1$ correspond to the extreme cases when there is no basis mixture and no dynamic adjustment, respectively. From Table \ref{Table: gate_forecasting_res_ode}, both cases lead to sub-optimal performance, especially when there is no modeling of dynamics, {\em i.e.}, $\gamma=1$. $\gamma=1e^{-2}$ is the optimal choice from grid search, which trades off the two cases for improved results. The choice of a small $\gamma$ also indicates a few introduction of basis mixture is sufficient. Moreover, the comparable results of cases (c) and (d) in Table \ref{Table: gate_forecasting_res_ode} validates the effective design of the gating function $\gamma(\cdot)$, which can help save a lot of tuning efforts.

\subsubsection{Visualization}
As discussed before, \ourmethod\ explores the latent clustering structures by learning a basis mixture distribution and tuning it over time. To understand how \ourmethod\ uncovers the clustering structures, we randomly sampled a batch of training (testing) samples for visualization (the full set is too large to be visualized). We visualized every temporal feature $\mat{x}_{t}$ using tSNE \cite{maaten2008visualizing} in a 2D space. The missing values in $\mat{x}_{t}$'s were imputed using the outputs of the forecasting component.

Fig. \ref{fig.vis} presents the visualization results and the Gaussian means $\boldsymbol{\mu}$ learned by \ourmethod-O (with gate function $\gamma(\cdot)$) on MIMIC-III and USHCN datasets. For the training set, we visualized all fitted data. For the testing set, we investigated the forecasting parts, {\em i.e.}, last 24 hours (20 days) on MIMIC-III (USHCN). In the figure, circles represent temporal features, ``+'' markers represent the learned Gaussian means. Different transparencies indicate different MTS samples. From the figure, we can clearly observe clusters. In particular, features $\mat{x}_{t}$ from different MTS's may stay in the same cluster, which implies different samples ({\em  e.g.}, patients) may share the same state ({\em  e.g.}, pathological condition) at different periods. From Fig. \ref{fig.vis}(a)(c), \ourmethod\ can effectively learn Gaussian means in dense areas from the training set, so that the mixture distributions are well fitted. Moreover, Fig. \ref{fig.vis}(b)(d) demonstrate the learned means fit the testing data as well, which explains how the new time series are forecasted meanwhile.

\section{Conclusions}

In this paper, we proposed a new method, Dynamic Gaussian Mixture based Deep Generative Model (\ourmethod), for robust forecasting on sparse multivariate time series (\mts). \ourmethod\ achieves robustness by modeling the transition of latent clusters of temporal features, and emitting MTS's from dynamic Gaussian mixture distributions. We parameterized the generative model by neural networks and developed an inference network for enabling inductive analysis. 
The extensive experimental results on a variety of real-life datasets demonstrated the effectiveness of our proposed method.


\clearpage
\section{Acknowledgments}
This material is based upon work that is in part supported by the Defense Advanced Research Projects Agency (DARPA) under Contract No. HR001117C0047.

\section{Ethical Impact Statement}

Multivariate time series forecasting has high impacts in wide domains, such as medicine, finance, and meteorology. In many emerging applications, MTS's collected from different sources are often interrelated and demands collective analysis for fully understanding the monitored targets, which necessitates solutions to handle sparsity. Such solutions do not only save engineering efforts for combining data, but also enhances predictive performance in important scenarios including clinical diagnosis, traffic surveillance, and large-scale system debugging. Our proposed model was tested on datasets from a variety of scenarios, which showcases some of its societal impacts.

Moreover, the theoretical design and analysis of the dynamic Gaussian mixture distribution itself may have broader impacts in applications other than MTS forecasting. Static Gaussian mixture has been extensively used in many applications, while its dynamic counterpart is less developed. Our proposed method provides a novel and general solution that explicitly defines temporal dependency between Gaussian mixture distributions at different time steps. It has a potential to be used on different types of sequential data such as sentences, dynamic graphs, and videos for modeling the flows of clustering structures which can be topics, social communities, and concepts, thus generates values in the correspondingly various areas.

\bibliography{ref}

\clearpage
\appendix



\section{Code availability}
The code of this paper can be found in the github repository: \url{https://github.com/thuwuyinjun/DGM2}.

\section{More Description on The Generative Process}

It is noteworthy that in our description of the generative process in Sec. ``Generative Model'', the starting latent variable is $z_{1}$, which is drawn from a uniform prior. From the prior, the predicted latent variables are $z_{2}$, ..., $z_{w+1}$  for $t=1, ... w$. Correspondingly, the generated sequence is $\mat{\tilde{x}}_{2}$, ..., $\mat{\tilde{x}}_{w+1}$, which was written in a way for highlighting the forecasting capability of the generative model.

Also, we want to clarify the consistency between the generative process and the objective function in Eq.~\eqref{eq.obj}.

For training our \ourmethod\ model, suppose a training sample has length $w$, {\em i.e.}, $\mat{x}_{1:w}$, which is used by the inference network to infer $z_{1}$, ..., $z_{w}$. As such, in the objective function in Eq.~\eqref{eq.obj}, we only used the generative model to generate $z_{2}$, ..., $z_{w}$ ({\em i.e.}, up to time step $w$, without $z_{w+1}$), 
to be in line with the inferred latent variables. Meanwhile, the prior $z_{1}$ is used for regularizing the inferred $z_{1}$ from the inference network for alleviating overfitting, and improving the generalization capability of the model. The forecasting capability of the model is preserved by the KL divergence in the second term on the transition probabilities.

Therefore, Eq.~\eqref{eq.obj} is consistent with the generative process, except for a notational difference on the last time step ({\em i.e.}, whether it is $w$ or $w+1$) of the input sequence.

\section{Derivation of Eq.~\eqref{eq.marginal_factor}}

In this section, we derive the factorized lower bound in Eq.~\eqref{eq.marginal_factor} as following.

\begin{proof}
First, according to the generative process in Sec. ``Generative Model'', the expectation of every generated temporal feature, {\em i.e.}, $\mat{\tilde{x}}_t$ can be written by
\begin{displaymath}
\small
\begin{aligned}
\small
\E(\mat{\tilde{x}}_t|z_{1:t-1}) &= \sum_{i=1}^k\E(\mat{\tilde{x}}_{t}|{\tilde{z}_{t}}=i)\p(\tilde{z}_{t} = i|z_{1:t-1})\\
& = \sum_{i=1}^k \E(\mat{\tilde{x}}_{t}|{\tilde{z}_{t}}=i)\boldsymbol{\psi}_{t}(\tilde{z}_{t} = i|z_{1:t-1}),
\end{aligned}
\end{displaymath}
where recall that $k$ is the number of mixture components, $z_t$ is the cluster variable for transition, $\tilde{z}_t$ is the cluster variable for emission, $\boldsymbol{\psi}_{t}$ is the dynamic Gaussian mixture distribution as described in Eq.~\eqref{eq.dynamic_probability} and $\E(\mat{\tilde{x}}_{t}|{\tilde{z}_{t}}=i)$ denotes the expectation of $\mat{\tilde{x}}_{t}$ when the cluster variable $\tilde{z}_{t}$ is $i$, i.e. the cluster centroid of the $i_{th}$ cluster. The above equation can be further decomposed into a dynamic adjustment part and a basis mixture part by substituting $\boldsymbol{\psi}_{t}$ with Eq.~\eqref{eq.dynamic_probability}. That is
\begin{align}\label{eq: exp_decomposition}
\small
    \begin{split}
    &\E(\mat{\tilde{x}}_t|z_{1:t-1})\\
&=\sum_{i=1}^k \E(\mat{\tilde{x}}_{t}|{\tilde{z}_{t}}=i)[(1-\gamma)p(z_{t}=i|z_{1:t-1}) + \gamma p(\boldsymbol{\mu_{i}})]\\
& = (1-\gamma)\sum_{i=1}^k \E(\mat{\tilde{x}}_{t}|{\tilde{z}_{t}}=i)p(z_{t}=i|z_{1:t-1})\\
&+ \gamma \sum_{i=1}^k \E(\mat{\tilde{x}}_{t}|{\tilde{z}_{t}}=i)p(\boldsymbol{\mu_{i}})\\
& = (1-\gamma)\sum_{i=1}^k \boldsymbol{\mu}_i p(z_{t}=i|z_{1:t-1}) + \gamma \sum_{i=1}^k \boldsymbol{\mu}_i p(\boldsymbol{\mu_{i}})
    \end{split}
\end{align}

The above decomposition implies that the emission step in the generation process can be regarded as a combination of two independent sub-emission processes, {\em i.e.}, one part of $\mat{\tilde{x}}_{t}$ is emitted from the dynamic adjustment while the other part is emitted from the basis mixture distribution. Therefore, the generative process can be rewritten to reflect this decomposition by:
\begin{outline}[enumerate]
\small
\1 for each MTS sample:
    \2 draw $z_{1} \sim \text{Uniform}(k)$
    \2 for time step $t=1, ..., w$:
        \3 compute transition probability by $p(z_{t+1}|z_{1:t})=f_{\boldsymbol{\theta}}(z_{1:t})$
        \3 draw $z_{t+1}\sim\text{Categorial}(p(z_{t+1}|z_{1:t}))$
        \3 draw a feature vector $\mat{\tilde{x}^{(1)}}_{t+1}\sim\mathcal{N}(\boldsymbol{\mu}_{{z}_{t+1}}, \sigma_1^{-1}\mathbf{I})$
        \3 draw 
        ${z}'_{t+1}\sim\text{Categorial}(p(\boldsymbol{\mu}))$
        \3 draw a feature vector $\mat{\tilde{x}}^{(2)}_{t+1}\sim\mathcal{N}(\boldsymbol{\mu}_{{z}'_{t+1}}, \sigma_2^{-1}\mathbf{I})$
        \3 output the feature vector $\mat{\tilde{x}}_{t + 1}' = (1-\gamma)\mat{\tilde{x}}^{(1)}_{t + 1} + \gamma\mat{\tilde{x}}^{(2)}_{t + 1}$
\end{outline}
where $\sigma_1$ and $\sigma_2$ satisfy $(1-\gamma)^2 \sigma_1^{-1} + \gamma^2 \sigma_2^{-1} = \sigma^{-1}$ such that:
\begin{align}\label{eq: distribution_combined}
\small
    \begin{split}
\mat{\tilde{x}}_{t + 1}' \sim \mathcal{N}((1-\gamma)\boldsymbol{\mu}_{{z}_{t+1}} + \gamma \boldsymbol{\mu}_{{z}'_{t+1}}, \sigma^{-1}\mathbf{I}).
    \end{split}
\end{align}

Therefore, the expectation of $\mat{\tilde{x}}_{t + 1}'$ given ${z}_{t+1}$ and ${z}'_{t+1}$ is:
\begin{displaymath}
\small
\begin{aligned}
& \E(\mat{\tilde{x}}_{t + 1}'|{z}_{t+1}, {z}_{t+1}') = \E(\mat{\tilde{x}}_{t + 1}'|{z}_{1:t+1}, {z}_{1:t+1}')\\
& = (1-\gamma)\boldsymbol{\mu}_{{z}_{t+1}} + \gamma \boldsymbol{\mu}_{{z}'_{t+1}}.
\end{aligned}
\end{displaymath}

Since $z_{t+1}\sim\text{Categorial}(p(z_{t+1}|z_{1:t}))$ and ${z}'_{t+1}\sim\text{Categorial}(p(\boldsymbol{\mu}))$, 
we can then compute the expectation of $\boldsymbol{\mu}_{{z}_{t+1}}$ and $\boldsymbol{\mu}_{{z}_{t+1}'}$ with respect to ${z}_{t+1}$ and ${z}_{t+1}'$, i.e.:
\begin{displaymath}
\small
\begin{aligned}
&\E_{z_{t+1}}\boldsymbol{\mu}_{{z}_{t+1}} = \sum_{z_{t+1}}\boldsymbol{\mu}_{{z}_{t+1}}p(z_{t+1}|z_{1:t}) = \sum_{i=1}^k \boldsymbol{\mu}_{i}p(z_{t+1}=i|z_{1:t})\\
& \E_{z_{t+1}'}\boldsymbol{\mu}_{{z}_{t+1}'} = \sum_{z_{t+1}'}\boldsymbol{\mu}_{{z}_{t+1}'}p(\boldsymbol{\mu}_{{z}_{t+1}'}) = \sum_{i=1}^k \boldsymbol{\mu}_{i}p(\boldsymbol{\mu}_{i}).
\end{aligned}
\end{displaymath}

Therefore, we can obtain the expectation of $\mat{\tilde{x}}_{t + 1}'$ by considering all the possible values of $z_{t+1}$ and $z_{t+1}'$ given the values of $z_{1:t}$, i.e.:
\begin{displaymath}
\small
\begin{aligned}
&\E(\mat{\tilde{x}}_{t + 1}'|{z}_{1:t}) = \sum_{z_{t+1}, z_{t+1}'}\E_{z_{t+1}, z_{t+1}'}(\mat{\tilde{x}}_{t + 1}'|z_{1:t+1}, z_{1:t+1}')\\
& = (1-\gamma)\E_{z_{t+1}}\boldsymbol{\mu}_{{z}_{t+1}} + \gamma \E_{z_{t+1}'}\boldsymbol{\mu}_{{z}'_{t+1}}\\
& = (1-\gamma)\sum_{i=1}^k \boldsymbol{\mu}_{i}p(z_{t+1}=i|z_{1:t}) + \gamma \sum_{i=1}^k \boldsymbol{\mu}_{i}p(\boldsymbol{\mu}_{i}).
\end{aligned}
\end{displaymath}

By replacing $t+1$ with $t$, the formula above becomes:
\begin{displaymath}
\small
\begin{aligned}
&\E(\mat{\tilde{x}}_{t}'|{z}_{1:t-1})= (1-\gamma)\E_{z_{t}}\boldsymbol{\mu}_{{z}_{t}} + \gamma \E_{z_{t}'}\boldsymbol{\mu}_{{z}'_{t}}\\
& = (1-\gamma)\sum_{i=1}^k \boldsymbol{\mu}_{i}p(z_{t}=i|z_{1:t-1}) + \gamma \sum_{i=1}^k \boldsymbol{\mu}_{i}p(\boldsymbol{\mu}_{i}),
\end{aligned}
\end{displaymath}

which is equivalent to the expectation of $\boldsymbol{\tilde{x}_t}$ as Eq.~\eqref{eq: exp_decomposition} indicates. Plus, $\boldsymbol{\tilde{x}_t}$ shares the same variance as $\boldsymbol{\tilde{x}_t}'$, i.e. $\sigma^{-1}\mathbf{I}$.
Therefore, $\boldsymbol{\tilde{x}_t}$ follows the same distribution as $\boldsymbol{\tilde{x}_t}'$, which indicates that the generative process presented above produces the same results as the one shown in Sec. ``Generative Model''. In what follows, the derivation will be based on this modified generative process.


Another by-product of the decomposition of the emission step in the generative process is that $z_t$ and $z_{t}'$ are two independent 
variables since they follow two independent distributions, $p(z_{t+1}|z_t)$ and $p(\boldsymbol{\mu})$, respectively. 


Given an input sequence $\mat{x}_{1:w}$, the goal is to maximize the log likelihood of this sequence, i.e. $p(\mat{x}_{1:w})$ , which requires to factorize $p(\mat{x}_{1:w})$ at each time step. To obtain the log likelihood of $\mat{x}_t$ at the time step $t$, we know that $\mat{x}_t$ follows the normal distribution as shown in Eq.~\eqref{eq: distribution_combined} and its log likelihood is:

\begin{align}\label{eq:density_deriv_0}
\small
\begin{split}
    &\log(p(\mat{x}_t | z_{t}, z_t'))\\
    &= \log(\sqrt{\frac{\sigma}{2 \pi}}e^{-\frac{1}{2}\sigma \|\mat{x}_t - (1-\gamma)\boldsymbol{\mu}_{{z}_{t}} - \gamma \boldsymbol{\mu}_{{z}'_{t}}\|^2})\\
    &= -\frac{1}{2}\sigma \|\mat{x}_t - (1-\gamma)\boldsymbol{\mu}_{{z}_{t}} - \gamma \boldsymbol{\mu}_{{z}'_{t}}\|^2  + C\\
    &= -\frac{1}{2}\sigma \|(1-\gamma)(\mat{x}_t - \boldsymbol{\mu}_{{z}_{t}}) + \gamma (\mat{x}_t- \boldsymbol{\mu}_{{z}'_{t}})\|^2  + C,
\end{split}
\end{align}
where $C=\log(\sqrt{\frac{\sigma}{2 \pi}})$ is a constant.


Using the convexity property of the function $f(\mat{x}) = \|\mat{x}\|^2$, and applying the Jensen's inequality, we can also decompose the formula above into the dynamic part and the basis mixture part, i.e.:
\begin{displaymath}
\small
\begin{aligned}
& \|(1-\gamma)(\mat{x}_t - \boldsymbol{\mu}_{{z}_{t}}) + \gamma (\mat{x}_t- \boldsymbol{\mu}_{{z}'_{t}})\|^2 \\
&\leq  (1-\gamma)\|\mat{x}_t - \boldsymbol{\mu}_{{z}_{t}}\|^2 + \gamma \|\mat{x}_t- \boldsymbol{\mu}_{{z}'_{t}}\|^2.
\end{aligned}
\end{displaymath}
Substituting the above inequality in Eq.~\eqref{eq:density_deriv_0}, we have: 
\begin{displaymath}
\small
\begin{aligned}
    & \log(p(\mat{x}_t | z_{t}, z_t'))\\
    & \geq -\frac{1}{2}\sigma\big[(1-\gamma)\|\mat{x}_t - \boldsymbol{\mu}_{{z}_{t}}\|^2 + \gamma \|\mat{x}_t- \boldsymbol{\mu}_{{z}'_{t}}\|^2\big] + C\\
    & = -\frac{1}{2}(1-\gamma)\sigma\|\mat{x}_t - \boldsymbol{\mu}_{{z}_{t}}\|^2 -\frac{1}{2}\gamma \sigma\|\mat{x}_t- \boldsymbol{\mu}_{{z}'_{t}}\|^2 + C\\
    & = (1-\gamma)(-\frac{1}{2}\sigma\|\mat{x}_t - \boldsymbol{\mu}_{{z}_{t}}\|^2 + C)\\
    & + \gamma (-\frac{1}{2}\sigma\|\mat{x}_t- \boldsymbol{\mu}_{{z}'_{t}}\|^2 + C),
\end{aligned}
\end{displaymath}
where $-\frac{1}{2}\sigma\|\mat{x}_t - \boldsymbol{\mu}_{{z}_{t}}\|^2 + C$ and $\frac{1}{2}\sigma\|\mat{x}_t- \boldsymbol{\mu}_{{z}'_{t}}\|^2 + C$ are equivalent to 
$p_{\boldsymbol{\vartheta}}(\mat{x}_t|z_t)$ and $p_{\boldsymbol{\vartheta}}(\mat{x}_t|z_t')$, respectively (i.e. the likelihood of $\mat{x}_t$ given $z_t$ and $z_t'$ respectively, parameterized by $\boldsymbol{\vartheta}$). 
Therefore, the above lower bound can be rewritten by
\begin{align}\label{eq: lower_bound_single_t}
\small
    \begin{split}
        & \log(p(\mat{x}_t | z_{t}, z_t')) \\
        & \geq (1-\gamma) \log(p_{\boldsymbol{\vartheta}}(\mat{x}_t|z_t)) + \gamma \log(p_{\boldsymbol{\vartheta}}(\mat{x}_t|z_t')).
    \end{split}
\end{align}



Next, we can derive the lower bound of $\log(p(\mat{x}_{1:w}))$. By introducing cluster variables $z_{1:w} = (z_1,z_2,\dots, z_w)$ and $z_{1:w}' = (z_1',z_2',\dots, z_w')$, 
and using the Jensen's inequality, while considering the fact that $\sum_{z_{1:w}, z_{1:w}'} p(z_{1:w}, z_{1:w}') = 1$, we have:
\begin{align}\label{eq: deriv0}
\small
\begin{split}
    &\log(p(\mat{x}_{1:w}))\\
    &= \log\big(\sum_{z_{1:w}, z_{1:w}'} p(\mat{x}_{1:w} | z_{1:w}, z_{1:w}')p(z_{1:w}, z_{1:w}')\big)\\
    &\geq \sum_{z_{1:w}, z_{1:w}'}\log\big(p(\mat{x}_{1:w} | z_{1:w}, z_{1:w}')\big) p(z_{1:w}, z_{1:w}').
\end{split}
\end{align}

Using the independence of $z_{1:w}$ and $z_{1:w}'$, and factorizing $p(\mat{x}_{1:w} | z_{1:w}, z_{1:w}')$ at each time step according to the emission step, Eq.~\eqref{eq: deriv0} can be further derived as
\begin{displaymath}
\small
\begin{aligned}
    &\log(p(\mat{x}_{1:w}))\\
    & \geq \sum_{z_{1:w}, z_{1:w}'}\log(p(\mat{x}_{1:w} | z_{1:w}, z_{1:w}')) p(z_{1:w}, z_{1:w}')\\
    & = \sum_{z_{1:w}, z_{1:w}'}\log(\Pi_{t=1}^w p(\mat{x}_{t} | z_{t}, z_{t}')) p_{\boldsymbol{\vartheta}}(z_{1:w}) p_{\boldsymbol{\vartheta}}(z_{1:w}')\\
    & = \sum_{z_{1:w}, z_{1:w}'} \sum_{t=1}^w \log(p(\mat{x}_{t} | z_{t}, z_{t}')) p_{\boldsymbol{\vartheta}}(z_{1:w}) p_{\boldsymbol{\vartheta}}(z_{1:w}'),
\end{aligned}
\end{displaymath}
where we explicitly incorporate the parameters $\boldsymbol{\vartheta} = \{\boldsymbol{\theta}, \boldsymbol{\mu}\}$ of $p(z_{1:w})$ and $p(z_{1:w}')$.

Then, by substituting $\log(p(\mat{x}_{t} | z_{t}, z_{t}'))$ with Eq.~\eqref{eq: lower_bound_single_t}, the above equation can be further bounded by
\begin{displaymath}
\small
\begin{aligned}
    &\log(p(\mat{x}_{1:w}))\\
    & \geq\sum_{z_{1:w}, z_{1:w}'} \sum_{t=1}^w \bigg[(1-\gamma) \log(p_{\boldsymbol{\vartheta}}(\mat{x}_t|z_t))\\
    &+ \gamma \log(p_{\boldsymbol{\vartheta}}(\mat{x}_t|z_t'))\bigg]p_{\boldsymbol{\vartheta}}(z_{1:w}) p_{\boldsymbol{\vartheta}}(z_{1:w}')\\
    & = \sum_{z_{1:w}, z_{1:w}'} \sum_{t=1}^w (1-\gamma) \log(p_{\boldsymbol{\vartheta}}(\mat{x}_t|z_t)) p_{\boldsymbol{\vartheta}}(z_{1:w}) p_{\boldsymbol{\vartheta}}(z_{1:w}')\\
    & + \sum_{z_{1:w}, z_{1:w}'} \sum_{t=1}^w \gamma \log(p_{\boldsymbol{\vartheta}}(\mat{x}_t|z_t')) p_{\boldsymbol{\vartheta}}(z_{1:w}) p_{\boldsymbol{\vartheta}}(z_{1:w}').
\end{aligned}
\end{displaymath}

Marginalizing $p(z_{1:w}')$ in the first term by summing up all values of $z_{1:w}'$, and marginalizing $p(z_{1:w})$ in the second term by summing up all values of $z_{1:w}$, we can get
\begin{align}\label{eq: prob_decomposition}
\small
\begin{split}
    &\log(p(\mat{x}_{1:w}))
    \geq (1-\gamma)\sum_{z_{1:w}} \sum_{t=1}^w \log(p_{\boldsymbol{\vartheta}}(\mat{x}_t|z_t)) p_{\boldsymbol{\vartheta}}(z_{1:w})\\
    &+ \gamma\sum_{z_{1:w}'} \sum_{t=1}^w \log(p_{\boldsymbol{\vartheta}}(\mat{x}_t|z_t'))p_{\boldsymbol{\vartheta}}(z_{1:w}'),
\end{split}
\end{align}
where $p_{\boldsymbol{\vartheta}}(z_{1:w})$ and $p_{\boldsymbol{\vartheta}}(z_{1:w}')$ can be factorized at each time step, respectively.
First, we factorize $p_{\boldsymbol{\vartheta}}(z_{1:w})$, {\em i.e.}, the joint probability in the dynamic adjustment part. Considering $z_t$ depends on $z_1, z_2,\dots, z_{t-1}$, we can sum up all the values of $z_{t+1}, z_{t+2},\dots, z_w$ in the first term of the lower bound in Eq.~\eqref{eq: prob_decomposition} at each time step by
\begin{align}\label{eq:deriv2}
\small
\begin{split}
    &\sum_{z_{1:w}} \sum_{t=1}^w \log(p_{\boldsymbol{\vartheta}}(\mat{x}_t|z_t)) p_{\boldsymbol{\vartheta}}(z_{1:w})\\
    &=\sum_{t=1}^w\sum_{z_{1:w}}\log(p_{\boldsymbol{\vartheta}}(\mat{x}_t|z_t)) p_{\boldsymbol{\vartheta}}(z_{1:w})\\
    & = \sum_{t=1}^w \sum_{z_{1:t}}\log( p_{\boldsymbol{\vartheta}}(\mat{x}_{t}|z_{t})) \sum_{z_{t+1:w}} p_{\boldsymbol{\vartheta}}(z_{1:w})  \\
    & = \sum_{t=1}^w \sum_{z_{1:t}}\log( p_{\boldsymbol{\vartheta}}(\mat{x}_{t}|z_{t})) p_{\boldsymbol{\vartheta}}(z_{1:t}) \\
    & = \sum_{t=1}^w \sum_{z_{1:t}}\log( p_{\boldsymbol{\vartheta}}(\mat{x}_{t}|z_{t})) p_{\boldsymbol{\vartheta}}(z_{t}| z_{1:t-1}) p_{\boldsymbol{\vartheta}}(z_{1:t-1})\\
    & = \sum_{t=1}^w \sum_{z_{1:t}}\log( p_{\boldsymbol{\vartheta}}(\mat{x}_{t}|z_{t})) p_{\boldsymbol{\theta}}(z_{t}| z_{1:t-1}) p_{\boldsymbol{\theta}}(z_{1:t-1}).
\end{split}
\end{align}

Next, we factorize $p_{\boldsymbol{\vartheta}}(z_{1:w}')$, {\em i.e.}, the joint probability in the basis mixture part. Since each $z_t$ is independent from the cluster variables at other time steps, the second term of the lower bound in Eq.~\eqref{eq: prob_decomposition} can be factorized at each time step by
\begin{align}\label{eq:deriv3}
\small
    \begin{split}
        &\sum_{z_{1:w}'} \sum_{t=1}^w \log(p_{\boldsymbol{\vartheta}}(\mat{x}_t|z_t'))p_{\boldsymbol{\vartheta}}(z_{1:w}')\\
        &=\sum_{t=1}^w \sum_{z_{1:w}'} \log(p_{\boldsymbol{\vartheta}}(\mat{x}_t|z_t'))p_{\boldsymbol{\vartheta}}(z_{1:w}')\\
        & = \sum_{t=1}^w \sum_{z_{t}'}\log( p_{\boldsymbol{\vartheta}}(\mat{x}_{t}|z_{t}')) \sum_{z_{1:t-1}', z_{t+1:w}'} p_{\boldsymbol{\vartheta}}(z_{1:w}')  \\
    & = \sum_{t=1}^w \sum_{z_t'}\log( p_{\boldsymbol{\vartheta}}(\mat{x}_{t}|z_{t}')) p_{\boldsymbol{\vartheta}}(z_t') \\
    & = \sum_{t=1}^w \sum_{z_{t}'}\log( p_{\boldsymbol{\vartheta}}(\mat{x}_{t}|z_{t}')) p(\boldsymbol{\mu}_{z_t'}).
    \end{split}
\end{align}

Since $z_t'$ and $z_t$ are independent in Eq.~\eqref{eq: prob_decomposition}, we can replace the notation $z_t'$ with $z_t$ in Eq.~\eqref{eq:deriv3}. Also, we can make use of the fact that $\sum_{z_{1:t-1}} p_{\boldsymbol{\theta}}(z_{1:t-1}) = 1$ to introduce an extra term in Eq.~\eqref{eq:deriv3} as
\begin{align}\label{eq:deriv4}
\small
    \begin{split}
        &\sum_{t=1}^w \sum_{z_{t}}\log( p_{\boldsymbol{\vartheta}}(\mat{x}_{t}|z_{t})) p(\boldsymbol{\mu}_{z_t}) \\
        & = \sum_{t=1}^w \sum_{z_{t}}\log( p_{\boldsymbol{\vartheta}}(\mat{x}_{t}|z_{t})) p(\boldsymbol{\mu}_{z_t}) \sum_{z_{1:t-1}} p_{\boldsymbol{\theta}}(z_{1:t-1})\\
        & = \sum_{t=1}^w \sum_{z_{1:t}}\log( p_{\boldsymbol{\vartheta}}(\mat{x}_{t}|z_{t})) p(\boldsymbol{\mu}_{z_t}) p_{\boldsymbol{\theta}}(z_{1:t-1})\\
    \end{split}
\end{align}

From Eq.~\eqref{eq: prob_decomposition}, Eq.~\eqref{eq:deriv2}, Eq.~\eqref{eq:deriv3}, and Eq.~\eqref{eq:deriv4}, we can obtain
\begin{displaymath}
\small
\begin{aligned}
& \log(p(\mat{x}_{1:w}))\\
& \geq (1-\gamma) \sum_{t=1}^w \sum_{z_{1:t}}\log( p_{\vartheta}(\mat{x}_{t}|z_{t})) p_{\boldsymbol{\theta}}(z_{t}| z_{1:t-1}) p_{\boldsymbol{\theta}}(z_{1:t-1}) \\
& + \gamma \sum_{t=1}^w \sum_{z_{1:t}}\log( p_{\vartheta}(\mat{x}_{t}|z_{t})) p(\boldsymbol{\mu}_{z_t}) p_{\boldsymbol{\theta}}(z_{1:t-1})\\
& = \sum_{t=1}^w \sum_{z_{1:t}}\bigg[\log( p_{\vartheta}(\mat{x}_{t}|z_{t})) p_{\boldsymbol{\theta}}(z_{1:t-1})\\ 
&~~~~~~~~~~~~~~~~~~~~[(1-\gamma) p_{\boldsymbol{\theta}}(z_{t}| z_{1:t-1}) + \gamma p(\boldsymbol{\mu}_{z_t})]\bigg],
\end{aligned}
\end{displaymath}
which completes the proof of Eq.~\eqref{eq.marginal_factor}.
\end{proof}

\section{Derivation of Eq. \eqref{eq.obj}}
In this section, we provide the derivation of the evidence lower bound (ELBO) in Eq.~\eqref{eq.obj}.
\begin{proof}
To address the computational difficulty of Eq.~\eqref{eq.marginal_factor}, we resort to variational inference, and introduce an approximated posterior probability $q_{\boldsymbol{\phi}}(z_{1:w}|\mat{x}_{1:w})$ into Eq.~\eqref{eq: deriv0} by
\begin{align}\label{eq: deriv5}
\small
\begin{split}
    & \log(p(\mat{x}_{1:w})) = \log( \sum_{z_{1:w}, z_{1:w}'} p(\mat{x}_{1:w} | z_{1:w}, z_{1:w}')p(z_{1:w}, z_{1:w}'))\\
    & = \log( \sum_{z_{1:w}, z_{1:w}'} p(\mat{x}_{1:w} | z_{1:w}, z_{1:w}')q_{\boldsymbol{\phi}}(z_{1:w}|\mat{x}_{1:w}) \frac{p(z_{1:w}, z_{1:w}')}{q_{\boldsymbol{\phi}}(z_{1:w}|\mat{x}_{1:w})})\\
    &= \log( \sum_{z_{1:w}, z_{1:w}'} p(\mat{x}_{1:w} | z_{1:w}, z_{1:w}')q_{\boldsymbol{\phi}}(z_{1:w}|\mat{x}_{1:w}) \frac{p(z_{1:w}) p(z_{1:w}')}{q_{\boldsymbol{\phi}}(z_{1:w}|\mat{x}_{1:w})})
\end{split}
\end{align}
where the last step is because of the independence of $z_{1:w}$ and $z_{1:w}'$.

Next, applying the Jensen's inequality based on $\sum_{z_{1:w}} q_{\boldsymbol{\phi}}(z_{1:w}|\mat{x}_{1:w}) = 1$ and $\sum_{z_{1:w}'} p(z_{1:w}') = 1$, and also factorizing $p(\mat{x}_{1:w} | z_{1:w}, z_{1:w}')$ at each time step, we have
\begin{displaymath}
\small
\begin{aligned}
    & \log(p(\mat{x}_{1:w})) \\
    & \geq \sum_{z_{1:w}, z_{1:w}'}\bigg[\log(p(\mat{x}_{1:w} | z_{1:w}, z_{1:w}')\frac{p(z_{1:w})}{q_{\boldsymbol{\phi}}(z_{1:w}|\mat{x}_{1:w})})\bigg]\\ 
    &~~~~~~~~~~~~~~~~~~~~\cdot q_{\boldsymbol{\phi}}(z_{1:w}|\mat{x}_{1:w}) p(z_{1:w}')\\
    & = \sum_{z_{1:w}, z_{1:w}'}\bigg[\log(\Pi_{t=1}^w p(\mat{x}_{t} | z_{t}, z_{t}')\frac{p(z_{1:w})}{q_{\boldsymbol{\phi}}(z_{1:w}|\mat{x}_{1:w})})\bigg]\\
    &~~~~~~~~~~~~~~~~~~~~\cdot q_{\boldsymbol{\phi}}(z_{1:w}|\mat{x}_{1:w})p(z_{1:w}')\\
    & = \sum_{z_{1:w}, z_{1:w}'}\bigg[\sum_{t=1}^w \log(p(\mat{x}_{t} | z_{t}, z_{t}')) + \log(\frac{p(z_{1:w})}{q_{\boldsymbol{\phi}}(z_{1:w}|\mat{x}_{1:w})})\bigg]\\
    &~~~~~~~~~~~~~~~~~~~~\cdot q_{\boldsymbol{\phi}}(z_{1:w}|\mat{x}_{1:w})p(z_{1:w}')\\
    & = \sum_{t=1}^w \sum_{z_{1:w}, z_{1:w}'} \log(p(\mat{x}_{t} | z_{t}, z_{t}'))q_{\boldsymbol{\phi}}(z_{1:w}|\mat{x}_{1:w})p(z_{1:w}')\\
    &~~~~~~~+ \sum_{z_{1:w}, z_{1:w}'} \log(\frac{p(z_{1:w})}{q_{\boldsymbol{\phi}}(z_{1:w}|\mat{x}_{1:w})})q_{\boldsymbol{\phi}}(z_{1:w}|\mat{x}_{1:w})p(z_{1:w}')
\end{aligned}
\end{displaymath}
where the second term can be simplified by summing up all the values of $z_{1:w}'$
\begin{displaymath}
\small
\begin{aligned}
    & \log(p(\mat{x}_{1:w})) \\
    & \geq \sum_{t=1}^w \sum_{z_{1:w}, z_{1:w}'} \log(p(\mat{x}_{t} | z_{t}, z_{t}'))q_{\boldsymbol{\phi}}(z_{1:w}|\mat{x}_{1:w})p_{\boldsymbol{\vartheta}}(z_{1:w}')\\
    & + \sum_{z_{1:w}} \log(\frac{p_{\boldsymbol{\vartheta}}(z_{1:w})}{q_{\boldsymbol{\phi}}(z_{1:w}|\mat{x}_{1:w})})q_{\boldsymbol{\phi}}(z_{1:w}|\mat{x}_{1:w}),
\end{aligned}
\end{displaymath}
where we incorporate the parameters $\mat{\boldsymbol{\vartheta}}$ of $p(z_{1:w})$ and $p(z_{1:w}')$.

Then, substituting $\log(p(\mat{x}_t | z_{t}, z_t'))$ with Eq.~\eqref{eq: lower_bound_single_t}, we have
\begin{displaymath}
\small
\begin{aligned}
    & \log(p(\mat{x}_{1:w})) \\
    & \geq \sum_{t=1}^w \sum_{z_{1:w}, z_{1:w}'} \bigg[(1-\gamma) \log(p_{\boldsymbol{\vartheta}}(\mat{x}_t|z_t)) + \gamma \log(p_{\boldsymbol{\vartheta}}(\mat{x}_t|z_t'))\bigg]\\
    & \cdot q_{\boldsymbol{\phi}}(z_{1:w}|\mat{x}_{1:w})p_{\boldsymbol{\vartheta}}(z_{1:w}')\\
    & + \sum_{z_{1:w}} \log(\frac{p_{\boldsymbol{\vartheta}}(z_{1:w})}{q_{\boldsymbol{\phi}}(z_{1:w}|\mat{x}_{1:w})})q_{\boldsymbol{\phi}}(z_{1:w}|\mat{x}_{1:w})\\
    & = (1-\gamma)\sum_{t=1}^w \sum_{z_{1:w}, z_{1:w}'} \log(p_{\boldsymbol{\vartheta}}(\mat{x}_t|z_t))q_{\boldsymbol{\phi}}(z_{1:w}|\mat{x}_{1:w})p_{\boldsymbol{\vartheta}}(z_{1:w}')\\
    & + \gamma \sum_{t=1}^w \sum_{z_{1:w}, z_{1:w}'} \log(p_{\boldsymbol{\vartheta}}(\mat{x}_t|z_t')) q_{\boldsymbol{\phi}}(z_{1:w}|\mat{x}_{1:w})p_{\boldsymbol{\vartheta}}(z_{1:w}')\\
    & + \sum_{z_{1:w}} \log(\frac{p_{\boldsymbol{\vartheta}}(z_{1:w})}{q_{\boldsymbol{\phi}}(z_{1:w}|\mat{x}_{1:w})})q_{\boldsymbol{\phi}}(z_{1:w}|\mat{x}_{1:w}).
\end{aligned}
\end{displaymath}

Summing up all the values of $z_{1:w}$ and $z_{1:w}'$ on the first term and the second term of the above equation respectively, we have:
\begin{align}\label{eq: prob_deriv1}
\small
\begin{split}
    & \log(p(\mat{x}_{1:w})) \\
    & \geq (1-\gamma)\underbrace{\sum_{t=1}^w \sum_{z_{1:w}} \log(p_{\boldsymbol{\vartheta}}(\mat{x}_t|z_t))q_{\boldsymbol{\phi}}(z_{1:w}|\mat{x}_{1:w})}_{\text{\objtermone}}\\
    & + \gamma \underbrace{\sum_{t=1}^w \sum_{z_{1:w}'} \log(p_{\boldsymbol{\vartheta}}(\mat{x}_t|z_t'))]p_{\boldsymbol{\vartheta}}(z_{1:w}')}_{\text{\objtermtwo}}\\
    & + \underbrace{\sum_{z_{1:w}} \log(\frac{p_{\boldsymbol{\vartheta}}(z_{1:w})}{q_{\boldsymbol{\phi}}(z_{1:w}|\mat{x}_{1:w})})q_{\boldsymbol{\phi}}(z_{1:w}|\mat{x}_{1:w})}_{\text{\objtermthree}}.
\end{split}
\end{align}

For the \textit{\objtermone}, we further sum up all the values of $z_{1:w}$ except $z_t$, then it is simplified as
\begin{align}\label{eq: obj_term1}
\small
    \begin{split}
        & \text{\objtermone} = \sum_{t=1}^w \sum_{z_t} \log(p_{\boldsymbol{\vartheta}}(\mat{x}_t|z_t))q_{\boldsymbol{\phi}}(z_t|\mat{x}_{1:t})\\
        & = \sum_{t=1}^w \E_{q_{\boldsymbol{\phi}}(z_t|\mat{x}_{1:t})} \log(p_{\boldsymbol{\vartheta}}(\mat{x}_t|z_t)).
    \end{split}
\end{align}

Similarly, for the \textit{\objtermtwo}, we sum up all the values of $z_{1:w}'$ except $z_t'$, it is simplified as
\begin{displaymath}
\small
\begin{aligned}
        & \text{\objtermtwo} = \sum_{t=1}^w \sum_{z_t'} \log(p_{\boldsymbol{\vartheta}}(\mat{x}_t|z_t'))p_{\boldsymbol{\vartheta}}(z_t')\\
        & = \sum_{t=1}^w \sum_{z_t'} \log(p_{\boldsymbol{\vartheta}}(\mat{x}_t|z_t'))p(\boldsymbol{\mu}_{z_t'}).
\end{aligned}
\end{displaymath}


Because $z_t$ and $z_t'$ are independent, we can replace $z_t'$ with $z_t$ so that the above equation can be rewritten by
\begin{align}\label{eq: obj_term2}
\small
    \begin{split}
        & \text{\objtermtwo} = \sum_{t=1}^w \sum_{z_t} \log(p_{\boldsymbol{\vartheta}}(\mat{x}_t|z_t))p(\boldsymbol{\mu}_{z_t}).
    \end{split}
\end{align}

For \textit{\objtermthree}, factorizing $p_{\boldsymbol{\vartheta}}(z_{1:w})$ and $q_{\boldsymbol{\phi}}(z_{1:w}|\mat{x}_{1:w})$ at each time step, we have
\begin{displaymath}
\small
\begin{aligned}
    &\text{\objtermthree}\\
    & = \sum_{z_{1:w}} \log(\frac{p(z_1)\Pi_{t=2}^w p_{\boldsymbol{\vartheta}}(z_{t}|z_{1:t-1})}{q_{\boldsymbol{\phi}}(z_{1}|\mat{x}_{1})\Pi_{t=2}^w q_{\boldsymbol{\phi}}(z_{t}|z_{t-1}, \mat{x}_{1:t})})q_{\boldsymbol{\phi}}(z_{1:w}|\mat{x}_{1:w})\\
    & = \sum_{z_{1:w}}\log(\frac{p(z_1)}{q_{\boldsymbol{\phi}}(z_{1}|\mat{x}_{1})}) q_{\boldsymbol{\phi}}(z_{1:w}|\mat{x}_{1:w})\\
    & + \sum_{t=2}^w \sum_{z_{1:w}} \log(\frac{p_{\boldsymbol{\vartheta}}(z_{t}|z_{1:t-1})}{q_{\boldsymbol{\phi}}(z_{t}|z_{t-1}, \mat{x}_{1:t})})q_{\boldsymbol{\phi}}(z_{1:w}|\mat{x}_{1:w}),
\end{aligned}
\end{displaymath}
where $p(z_1)$ is the uniform prior.

Summing up all the values of the cluster variables $z_{t+1},z_{t+2},\dots, z_w$ for each summand term, the above equation becomes
\begin{align}\label{eq: obj_term3}
\small
    \begin{split}
        &\text{\objtermthree}\\
    & = \sum_{z_{1}}\log(\frac{p(z_1)}{q_{\boldsymbol{\phi}}(z_{1}|\mat{x}_{1})}) q_{\boldsymbol{\phi}}(z_{1}|\mat{x}_{1})\\
    & + \sum_{t=2}^w \sum_{z_{1:t}} \log(\frac{p_{\boldsymbol{\vartheta}}(z_{t}|z_{1:t-1})}{q_{\boldsymbol{\phi}}(z_{t}|z_{t-1}, \mat{x}_{1:t})})q_{\boldsymbol{\phi}}(z_{1:t}|\mat{x}_{1:t})\\
    & = -\mathcal{D}_{KL} (q_{\boldsymbol{\phi}}(z_{1}|\mat{x}_{1}) || p(z_1)) \\
    & + \sum_{t=2}^w \sum_{z_{1:t}} \log(\frac{p_{\boldsymbol{\vartheta}}(z_{t}|z_{1:t-1})}{q_{\boldsymbol{\phi}}(z_{t}|z_{t-1}, \mat{x}_{1:t})})q_{\boldsymbol{\phi}}(z_{t}|\mat{x}_{1:t}, z_{t-1})q_{\boldsymbol{\phi}}(z_{1:t-1}|\mat{x}_{1:t-1})\\
    & = -\mathcal{D}_{KL} (q_{\boldsymbol{\phi}}(z_{1}|\mat{x}_{1}) || p(z_1)) \\
    & + \sum_{t=2}^w -\E_{q_{\boldsymbol{\phi}}(z_{1:t-1}|\mat{x}_{1:t-1})}[\mathcal{D}_{KL}(q_{\boldsymbol{\phi}}(z_{1:t}|\mat{x}_{1:t}) || p_{\boldsymbol{\vartheta}}(z_{t}|z_{1:t-1}))]\\
    & = -\mathcal{D}_{KL} (q_{\boldsymbol{\phi}}(z_{1}|\mat{x}_{1}) || p(z_1)) \\
    & -\sum_{t=1}^{w-1} \E_{q_{\boldsymbol{\phi}}(z_{1:t}|\mat{x}_{1:t})}[\mathcal{D}_{KL}(q_{\boldsymbol{\phi}}(z_{1:t+1}|\mat{x}_{1:t+1}) || p_{\boldsymbol{\vartheta}}(z_{t+1}|z_{1:t}))]
    \end{split}
\end{align}

From Eq.~\eqref{eq: obj_term1}-\eqref{eq: obj_term3}, Eq.~\eqref{eq: prob_deriv1} can be rewritten by

\begin{displaymath}
\small
\begin{aligned}
    & \log(p(\mat{x}_{1:w})) \\
    & \geq \text{\objtermone} + \text{\objtermtwo} + \text{\objtermthree}\\
    & \geq (1-\gamma)\sum_{t=1}^w \E_{q_{\boldsymbol{\phi}}(z_{t}|\mat{x}_{1:t})}\log (p_{\vartheta}(\mat{x}_{t}|z_{t})) \\
    & -\mathcal{D}_{KL}(q_{\boldsymbol{\phi}}(z_{1}|\mat{x}_{1}) || p(z_1)) \\
    & - \sum_{t=1}^{w-1}\E_{q_{\boldsymbol{\phi}}(z_{1:t}|\mat{x}_{1:t})}[\mathcal{D}_{KL}(q_{\boldsymbol{\phi}}(z_{t+1}|\mat{x}_{1:t+1}, z_t) || p_{\boldsymbol{\vartheta}}(z_{t+1}|z_{1:t}))] \\
    & + \gamma \sum_{t=1}^w \sum_{z_{t}}\log( p_{\vartheta}(\mat{x}_{t}|z_{t})) p(\boldsymbol{\mu}_{z_t}),
\end{aligned}
\end{displaymath}
which completes the proof of Eq.~\eqref{eq.obj}.
\end{proof}

\section{The Analytical Form of Eq. \eqref{eq.obj}}

In this section, we provide the analytical form of Eq.~\eqref{eq.obj}. As mentioned in the generative process in the subsection ``Forecasting Component'', $\mat{x}_{t}\sim\mathcal{N}(\boldsymbol{\mu}_{z_{t}}, \sigma^{-1}\mathbf{I})$. Therefore, the term $\log(p_{\vartheta}(\mat{x}_{t}|z_{t}))$ can be written by
\begin{align}\label{eq: analytic_1}
\small
    \begin{split}
\log(p_{\vartheta}(\mat{x}_{t}|z_{t})) = -\frac{1}{2}\sigma||\mat{x}_t - \boldsymbol{\mu}_{z_{t}}||^2 + \log(\sqrt{\frac{\sigma}{2\pi}})        
    \end{split}
\end{align}

Also, the KL divergence between two distributions, $\mathcal{D}_{KL}(q_{\boldsymbol{\phi}}(z_{t+1}|\mat{x}_{1:t+1}, z_t) || p_{\boldsymbol{\vartheta}}(z_{t+1}|z_{1:t}))$, can be written by (recall that $k$ is the number of clusters)
\begin{align}\label{eq: analytic_2}
\small
\begin{split}
    &\mathcal{D}_{KL}(q_{\boldsymbol{\phi}}(z_{t+1}|\mat{x}_{1:t+1}, z_t) || p_{\boldsymbol{\vartheta}}(z_{t+1}|z_{1:t}))\\
    & = \sum_{r=1}^k q_{\boldsymbol{\phi}}(z_{t+1}=r|\mat{x}_{1:t+1}, z_t) \log(\frac{q_{\boldsymbol{\phi}}(z_{t+1}=r|\mat{x}_{1:t+1}, z_t)}{p_{\boldsymbol{\vartheta}}(z_{t+1}=r|z_{1:t})})
\end{split}    
\end{align}

Therefore, from Eq. \eqref{eq: analytic_1} and Eq. \eqref{eq: analytic_2}, the analytical form of Eq. \eqref{eq.obj} is

\begin{align}
\small
    \begin{split}
    & \ell(\boldsymbol{\vartheta}, \boldsymbol{\phi})\\
    & = (1-\gamma)\sum_{t=1}^w \sum_{r=1}^k q_{\boldsymbol{\phi}}(z_{t}=r|\mat{x}_{1:t})[-\frac{1}{2}\sigma||\mat{x}_t - \boldsymbol{\mu}_{r}||^2 + \log(\sqrt{\frac{\sigma}{2\pi}})]\\
    &- \sum_{r=1}^k q_{\boldsymbol{\phi}}(z_{1}=r|\mat{x}_{1}) \log(\frac{q_{\boldsymbol{\phi}}(z_{1}=r|\mat{x}_{1})}{p_{\boldsymbol{\vartheta}}(z_{1}=r)})\\
    & - \sum_{t=1}^{w-1}\sum_{r=1}^k\left[\E_{q_{\boldsymbol{\phi}}(z_{1:t}|\mat{x}_{1:t})} q_{\boldsymbol{\phi}}(z_{t+1}=r|\mat{x}_{1:t+1}, z_t)\right.\\
    &\left.\cdot\log(\frac{q_{\boldsymbol{\phi}}(z_{t+1}=r|\mat{x}_{1:t+1}, z_t)}{p_{\boldsymbol{\vartheta}}(z_{t+1}=r|z_{1:t})})\right]\\
    & + \gamma \sum_{t=1}^w \sum_{r=1}^k p(\boldsymbol{\mu}_{z_t}=r)\left[-\frac{1}{2}\sigma||\mat{x}_t - \boldsymbol{\mu}_{r}||^2 + \log(\sqrt{\frac{\sigma}{2\pi}})\right]
    \end{split}
\end{align}

\section{Subroutine to Evaluate The Probability $\q(\z_t|\bx_{1:t})$ in Eq. \eqref{eq.obj}}

As mentioned before, the joint probability $\q(\z_t|\bx_{1:t})$ is essential for the evaluations of the first term in Eq. \eqref{eq.obj}, which, however, is unable to be directly attained through the inference network. As a consequence, Algorithm \ref{alg: marginal_prob_alg} is used for evaluating this marginal probability recursively by utilizing the conditional probability $\q(\z_{t+1}| \z_{t}, \bx_{1:t+1})$ produced by the inference network. 

\begin{algorithm}
\small
\SetAlgoLined
\KwResult{$\q(\z_t|\bx_{1:t})$ at every time step $t$}

 \For{$t = 1 ; t <= T-1; t++$}{
$\q(\z_{t+1} = r|\bx_{1:t+1})$

$= \sum_{s=1}^k \q(\z_{t+1} = r | \z_{t}=s, \bx_{1:t+1})\q(\z_{t}=s | \bx_{1:t+1})$

$= \sum_{s=1}^k \q(\z_{t+1} = r | \z_{t}=s, \bx_{1:t+1})\q(\z_{t}=s | \bx_{1:t})$

$(r=1,2,\dots, k)$
  
  }
 \caption{Subroutine to compute $\q(\z_t|\bx_{1:t})$}
 \label{alg: marginal_prob_alg}
\end{algorithm}

\section{Integrating ODE Units in \ourmethod}
Ordinary differential equation unit (ODE) is recently proposed in \cite{chen2018neural} as a way to model the dynamics of the continuous-time hidden states by leveraging the following ordinary equations on the hidden states with respect to time:
$$\frac{d \mat{h}(t)}{d t} = f(\mat{h}(t), t, \theta)$$

in which $\mat{h}(t)$ represents the hidden states at time step $t$ and the function $f$ is a parameterized function to compute the gradient of the hidden states.
Then given a hidden state at the initial time step $t_0$, i.e. $\mat{h}_0$, the hidden state at arbitrary time step $t$ can be evaluated by using a numeric ODE solver, i.e.:
\begin{equation}\label{eq: ode_solve}
\mat{h}(t) = \text{ODESolve}(f(\mat{h}_0, (t_0, t)))    
\end{equation}

Based on the generative process above, \cite{rubanova2019latent} designed a variational autoencoder structure in which given the initial hidden state $\mat{h}_0$, the hidden states at the time step $t$ are computed by using Eq.~\eqref{eq: ode_solve} in the generative process, while the hidden state $\mat{h}_t$ at the time step $t$ can also be inferred by a hybrid structure called \odernn\ in the inference process, which employs ODE 
to model the dynamics between two consecutive observations and leverage RNN to update the hidden states at each observation, i.e.:
\begin{equation}\label{eq.ode_rnn}
\small
\begin{aligned}
& \mat{h}_i' = \text{ODESolve}(f(\tilde{\mat{h}}_{i-1}, (t_{i-1}, t_i)))\\
& \tilde{\mat{h}}_i = \text{RNN}(\mat{h}_i', \bx_i)\\
\end{aligned}
\end{equation}

 
 
 
 
 





To integrate the ODE units into our framework, we applied the following adjustments to both the generative network and the inference network respectively.


\subsection{Integrating ODE into the generative network}

By comparing the standard recurrent neural network architecture against Eq.~\eqref{eq.ode_rnn}, the only difference is the additional equation, i.e. the first equation in Eq.~\eqref{eq.ode_rnn} to encode the continuous time intervals between two consecutive observations. Therefore, to model the dynamics of the clustering variables in the generative network with ODE units, we can substitute the RNN term in Eq.~\eqref{eq.gen_rnn} with Eq.~\eqref{eq.ode_rnn} 
when the continuous-time intervals exist, i.e.:
\begin{equation}\label{eq.gen_ode}
\small
\begin{aligned}
&p(z_{i+1}|z_{1:i})=\text{softmax}(\text{MLP}({\mat{h}}_{i+1}))\\
& \text{where}~~{\mat{h}}_{i+1} = \text{ODESolve}(f(\mat{h}_{i}', (t_{i}, t_{i+1}))) \\
& \text{and}~~\mat{h}_{i}' = \text{RNN}({\mat{h}}_{i}, \z_i)\\
\end{aligned}
\end{equation}

Note that compared to Eq.~\eqref{eq.gen_rnn}, the formula above is enhanced by taking the time intervals into consideration, in which $\mat{h}_{i}'$ in the formula above plays the same role as $\mat{h}_t$ (the output of RNN) but the combination of $\mat{h}_{i}'$ and the encoded time interval rather than $\mat{h}_{i}'$ itself is used for generating $z_t$. To facilitate the forecasting of the observations at an arbitrary future time step $t^{*} (t^* > t_w)$, we slightly modify the formulas by employing the hidden states at the last observed time step $w$ for computing the hidden states in the future time steps, i.e.:
\begin{equation}\label{eq.forecasting_ode}
\small
\begin{aligned}
&p(z^*|z_{1:w})=\text{softmax}(\text{MLP}({\mat{h}}^{*}))\\
&\text{where}~~{\mat{h}}^{*} = \text{ODESolve}(f(\mat{h}_w', (t_{w}, t^*)))\\
& \text{and}~~\mat{h}_w' = \text{RNN}({\mat{h}}_w, \z_w)\\
\end{aligned}
\end{equation}

The equation above builds direct dependency between the last observed time step $t_w$ and the arbitrary future time step $t^*$, which is expected to lead to more accurate forecasting results than \ourmethod-L. Since in \ourmethod-L, the forecasting results at time step $t^*$ is determined by the time step $t^* - 1$ (rather than $t_w$) where the forecasting results already deviate from the expected observations, the errors are thus accumulated for forecasting the far future time steps, leading to low forecasting accuracy in those time steps, which, however can be mitigated if forecasting the future time steps is based on Eq.~\eqref{eq.forecasting_ode}.



\subsection{Integrate ODE into the inference network}

To integrate ODE into the inference network, we use ODE-RNN (i.e. Eq.~\eqref{eq.ode_rnn}) instead of RNN to obtain the hidden representation $\tilde{\mat{h}}_t$ of the observation $\bx_{t}$, which are then fed into the MLP and softmax layers for generating the approximate posterior probability $q_{\boldsymbol{\phi}}(z_{i+1}|\mat{x}_{1:i+1}, z_{i})$, i.e.:
\begin{equation}\label{eq.infer_ode}
\small
\begin{aligned}
&q_{\boldsymbol{\phi}}(z_{i+1}|\mat{x}_{1:i+1}, z_{i})=\text{softmax}(\text{MLP}(\mat{\tilde{h}}_{i+1}))\\
& \text{where}~~\mat{\tilde{h}}_{i+1} = \text{RNN}(\mat{h}_{i+1}', \bx_{i+1})\\
&\text{and}~~\mat{h}_{i+1}' = \text{ODESolve}(f(\mat{{\tilde{h}}}_{i}, (t_{i}, t_{i+1})))\\
\end{aligned}
\end{equation}

\subsection{Capture long-term dependency}
As indicated above, the generation of $z_{i+1}$ in the generative network integrated with ODEs still depends on all the previous states $z_{1:i}$, which is, however, insufficient to construct accurate forecasting results at the far future time steps. To deal with this issue, we leverage the property of ODE that it can construct the direct dependency between hidden states with longer continuous time intervals. Specifically, we generate $z_i$ with $z_{1:i-r}$ where $r$ can be arbitrary integer, which is expressed as:

\begin{equation}\label{eq.gen_ode2}
\small
\begin{aligned}
&p(\z_{i+1}|\z_{1:i-r+1})=\text{softmax}(\text{MLP}({\mat{h}}_{i+1}))\\
&\text{and}~~{\mat{h}}_{i+1} = \text{ODESolve}(f(\mat{h}_{i-r+1}', (t_{i-r+1}, t_{i+1})))\\
&\text{where}~~\mat{h}_{i-r+1}' = \text{RNN}({\mat{h}}_{i-r+1}, \z_{i-r+1})
\end{aligned}
\end{equation}

in which $\z_{i+1}$ directly depends on the cluster variables in the first $i-r+1$ time steps, i.e. $\z_{1:i-r+1}$. 
The resulting distribution $p(\z_{i+1}|\z_{1:i-r+1})$ in Eq.~\eqref{eq.gen_ode2} is also supposed to fit the categorical distribution $q_{\boldsymbol{\phi}}(z_{i+1}|\mat{x}_{1:i+1}, z_{i})$ inferred by the inference network. Therefore, we expect to minimize the KL-divergence between $p(\z_{i+1}|\z_{1:i-r+1})$ and $q_{\boldsymbol{\phi}}(z_{i+1}|\mat{x}_{1:i+1}, z_{i})$, which requires one additional term in the ELBO formula shown in Eq.~\eqref{eq.obj} (weighted by a hyper-parameter $\lambda$), i.e.:

\begin{displaymath}\label{eq.obj2}
\small
\begin{aligned}
&\ell(\boldsymbol{\vartheta}, \boldsymbol{\phi}) = \gamma\sum_{i=1}^{w}\E_{q_{\boldsymbol{\phi}}(\z_i|\mat{x}_{1:i})}[\log{(p_{\boldsymbol{\vartheta}}(\bx_i|\z_i))}]\\
&- \sum_{i=1}^{w-1}\E_{q_{\boldsymbol{\phi}}(z_{1:i}|\mat{x}_{1:i})}[\mathcal{D}_{KL}\big(q_{\boldsymbol{\phi}}(z_{i+1}|\mat{x}_{1:i+1}, z_{i}) || p_{\boldsymbol{\vartheta}}(\z_{i+1}|\z_{1:i})\big)]\\
&- \lambda\underbrace{\sum_{i=r+1}^{w}\E_{q_{\boldsymbol{\phi}}(z_{1:i-r}|\mat{x}_{1:i-r})}[\mathcal{D}_{KL}\big(q_{\boldsymbol{\phi}}(z_{i}|\mat{x}_{1:i}, z_{i-1}) || p_{\boldsymbol{\vartheta}}(\z_{i}|\z_{1:i-r})\big)]}_{\text{Additional KL divergence}}\\
&- \mathcal{D}_{KL}\big(q_{\boldsymbol{\phi}}(\z_{1}|\mat{x}_{1}) || p_{\boldsymbol{\vartheta}}(z_1)\big)\\
&+ (1-\gamma)\sum_{t=1}^{w}\sum_{z_{i}=1}^{k}p(\boldsymbol{\mu}_{z_{i}})\log{(p_{\boldsymbol{\vartheta}}(\mat{x}_{i}|z_{i}))}
\end{aligned}
\end{displaymath}

By comparing Eq.~\eqref{eq.gen_ode} and the formula above, we notice that the generation of $z_{t+1}$ depends on cluster variables in the far earlier time steps. Therefore, intuitively, the minimization of the additional KL divergence can aid maintaining the dynamics of the hidden states in the longer period, thus resulting in better forecasting performance. In the experiments, $r$ was searched within $\{5,10,15,20\}$.

\section{The Configurations of The Compared Methods in The Experiments}
Similar to \ourmethod, the hidden dimensionality in the RNNs (LSTM, GRU or ODE) used in the compared methods was searched within $\{10, 20, 30, 40, 50\}$. 
For the methods using variational inference ({\em e.g.}, \dmm\ and \lode), the variance of the Gaussian distribution was searched within $\{1e^{-5}, 1e^{-4}, 1e^{-3},1e^{-2},1e^{-1}\}$. Note that not all of the compared methods were originally designed for the forecasting tasks, {\em e.g.}, \lstm\ and \grud, for which we added an MLP layer to map the feature vectors or the hidden representations at each time step to a predicted temporal feature. This MLP contains one hidden layer with dimensionality selected from $\{10,20, 30, 40, 50\}$. Other model-specific configurations are described in the following.

For \var, the future observation $\mat{x}_{t+1}$ is assumed to be linearly dependent on the last $p$ observations i.e., $\mat{x}_{t-p+1:t}$ where $p$ was searched within $[1, 20]$.

For \grui, the number of epochs for the imputation step used by the adversarial learning was selected from 0 to 100 such that the convergence of the imputation step is reached.

For \ipnet, the reference time steps of the input \mts\ were set as $\{t_1',t_2',\dots, t_w'\} = \{1,2,\dots, w\}$. 



For \xgboost, the xgboost package from the scikit-learn library\footnote{https://scikit-learn.org/stable/} was used, where the number of gradient boosted trees was searched from 1 to 5.

\section{Parameter sensitivity evaluation}

\begin{table}[!t]
\centering
\small
\caption{Analysis of hyper-parameter $k$}\label{Table: forecasting_varied_k}
\vspace*{-0.2cm}
\begin{tabular}{c|ccccc} \hline
$k$ &10 & 20 & 50 & 100 & 200  \\ \hline
RMSE& 0.5868 & 0.5450 & 0.5426 & 0.5430 & 0.5432  \\ 
MAE & 0.4299 & 0.3901 & 0.3848 & 0.3849 & 0.3855  \\ \hline
\end{tabular}
\end{table}

We also studied the sensitivity of the hyper-parameters in \ourmethod, including the number of clusters $k$, and the strength of the basis mixture component $\gamma$. Note that since the effect of $\gamma$ has been investigated in the Sec. ``Experiments' (see Table \ref{Table: gate_forecasting_res_ode}), only the effect of $k$ is explored in this section. In this experiment, we measured the forecasting performance of 5 runs on \climate\ datasets by using \ourmethod-L with different values of $k$ and reported average forecasting errors (\rmse\ and \mae) in Table \ref{Table: forecasting_varied_k}. Similar performance trends are also observed in other datasets, which are thus omitted here. 

According to Table \ref{Table: forecasting_varied_k}, we can know that the value of $k$ can influence the performance of \ourmethod\ and 
with a reasonable value of $k$ (e.g. 50 for \climate\ dataset), \ourmethod\ can achieve the best forecasting performance. When $k$ is too small, e.g. 10 for \climate\ dataset, the produced cluster centroids fail to cover all the clusters in the dataset, thereby hurting the forecasting performance. In contrast, with very large $k$, we can expect that a large portion of the produced cluster centroids are overlapped, resulting in slightly higher forecasting errors due to overfitting, and longer training time. Hence, we picked up the value of $k$ between 20 and 100 to balance 
the computational overhead and the forecasting accuracy.

\end{document}



